%% file: main_arxiv_v1.tex
\documentclass{article} 
\usepackage{arxiv_tmpl,times}
\usepackage{xcolor}
\usepackage{simpleicons}
\input{math_commands.tex}

\usepackage{pifont}
\usepackage{capt-of}
\usepackage{algorithm}
\usepackage{algpseudocode}
\usepackage{amsmath}
\usepackage{etoc}
\usepackage{amsthm}
\usepackage{hyperref}
\usepackage{url}
\usepackage{graphicx}
\usepackage{booktabs}   
\usepackage{array}      
\usepackage{multirow}   
\usepackage[table]{xcolor}
\usepackage{cleveref}  
\usepackage[margin=1in]{geometry}
\usepackage{xcolor}
\usepackage[most]{tcolorbox}
\usepackage{enumitem}
\usepackage{longtable}
\usepackage{amsthm}
\usepackage[table]{xcolor}
\definecolor{WorseRed}{HTML}{C0392B}
\definecolor{NoiseGrey}{HTML}{7F8C8D}
\definecolor{BetterGreen}{HTML}{1E8449}
\definecolor{adaevacolor}{HTML}{ffefe8}

\definecolor{huggingfaceyellow}{HTML}{FFD21E}
\usepackage{fontawesome5}

\usepackage{graphicx,framed}
\definecolor{MyBlue}{rgb}{0.25,0.5,0.75}
\colorlet{shadecolor}{MyBlue!15}
{\begin{snugshade}\begin{quote}}
{\hfill\end{quote}\end{snugshade}}

\title{AdaEva: Accelerating LLM-Driven Algorithm Design with Adaptive Partial Evaluation}

\author{%
Tai Nguyen\textsuperscript{1,2}\thanks{Corresponding author: \texttt{dptn1@st-andrews.ac.uk}} \quad
Fei Liu\textsuperscript{3} \quad
Phong Le\textsuperscript{1} \quad
Carola Doerr\textsuperscript{2} \quad
Nguyen Dang\textsuperscript{1}
\\
\textsuperscript{1}University of St Andrews, United Kingdom
\\
\textsuperscript{2}Sorbonne Universit{\'e}, CNRS, LIP6, France
\\
\textsuperscript{3}University of Zurich and ETH Zurich, Switzerland
}

\definecolor{mistralorange}{HTML}{FF7000}
\definecolor{mistralbg}{HTML}{FFF4EC}
\tcbset{
  redpromptbox/.style={
    enhanced,
    colback=mistralbg,
    colframe=mistralorange,
    boxrule=0.5pt,
    leftrule=3.5pt,
    rightrule=3.5pt,
    sharp corners,
    left=12pt,
    right=12pt,
    top=8pt,
    bottom=8pt,
    coltitle=black,
    fonttitle=\bfseries\normalsize,
    attach title to upper=\par\vspace{6pt}
  }
}

\iclrfinalcopy 
\begin{document}

\maketitle

\begin{abstract}

Large Language Models (LLMs) are increasingly used for automated algorithm design. However the computational cost of evaluating the generated algorithms can be excessive. We consider the common LLM-driven automated algorithm design (LLM4AD) setting in which a candidate algorithm is evaluated by aggregating its performance over a shared set of training instances. This instance-wise structure raises a natural question: \emph{must every candidate be evaluated on the entire instance set before deciding whether it remains competitive?} Taking inspiration from algorithm configuration, we introduce AdaEva, a drop-in \textbf{ada}ptive partial-\textbf{eva}luation framework that progressively evaluates candidates on larger subsets of the same instance pool and eliminates unpromising candidates as evidence accumulates. Importantly, AdaEva leaves the underlying LLM4AD procedure and per-instance evaluator unchanged and requires no prior knowledge about instance difficulty. We instantiate this idea using successive halving (AdaEva-S) and statistical racing (AdaEva-R), and evaluate both mechanisms across three representative LLM4AD frameworks, multiple LLM backbones, and optimization domains spanning combinatorial and continuous black-box optimization. Under matched evaluation budgets, AdaEva more reliably balances evaluation effort across candidates than fixed partial-evaluation strategies, yielding strong search efficiency and anytime performance together with improved held-out generalization across the evaluated settings. 

\end{abstract}

\begin{center}
\small
\textbf{Code and Experimental Materials:}
\href{https://github.com/taindp98/AdaEva}{\faGithub\ AdaEva}
\end{center}

\addtocontents{toc}{\protect\setcounter{tocdepth}{-1}}
\section{Introduction}
\label{introduction}

Recent advances in Large Language Models (LLMs) have generated substantial interest in LLM-driven automated algorithm design (LLM4AD)~\citep{liu2026systematic}, where LLMs are used to generate and iteratively refine complex heuristic algorithms for challenging optimization tasks~\citep{romera2024mathematical, liu2024eoh, van2025llamea, ye2024reevo}. While most work focuses on improving generated algorithms or reducing LLM inference cost, another major source of computational overhead has received comparatively little attention: \emph{candidate evaluation}.

In many LLM4AD applications, evaluating a generated algorithm requires executing it repeatedly over a collection of problem instances. In combinatorial optimization, for example, each candidate heuristic may need to solve tens or hundreds of training instances before its quality can be estimated reliably. Similar costs arise in continuous black-box optimization, where candidate algorithms are typically evaluated across multiple functions, dimensions, or random seeds. As the number of generated candidates increases, exhaustive evaluation can therefore consume a substantial fraction of the overall computational budget, limiting the number of algorithms that can be explored.
This motivates the central question of our work:
\emph{Can we reduce candidate-evaluation cost in evolutionary LLM4AD without sacrificing the quality of the algorithms discovered?}

We study the common LLM4AD setting in which candidate quality is measured by aggregating performance over a set of training instances. This structure, arising naturally in heuristic and algorithm design for optimization problems, creates an opportunity for \emph{adaptive evaluation}. Rather than evaluating every candidate immediately on the full instance set, we can first evaluate candidate algorithms on a small subset and allocate additional evaluations only to those that seem promising. Clearly inferior candidates can thus be discarded early, before consuming the full evaluation budget.

Building on this observation, we introduce \textbf{AdaEva}, an \textbf{Ada}ptive \textbf{Eva}luation framework for LLM-driven algorithm design. AdaEva acts as an evaluation layer on top of an existing LLM4AD search process (see Figure~\ref{fig:AdaEva}): candidate generation, prompting, evolutionary operators, and the underlying problem evaluator remain unchanged, while evaluation effort is allocated adaptively based on the evidence accumulated during the search. AdaEva progressively allocates additional evaluations to candidates that remain competitive while eliminating unpromising candidates early, so that adaptive resource allocation is done without modifying the underlying LLM4AD framework's search mechanism.

This general design naturally accommodates resource-allocation techniques developed in the adjacent areas of automated algorithm configuration~\citep{hoos2012automated} and hyper-parameter optimization~\citep{yang2020hyperparameter}. We instantiate AdaEva using two complementary approaches. The first, \textbf{AdaEva-S}, is based on \emph{\underline{s}uccesive halving}~\citep{karnin2013almost}, which progressively allocates larger evaluation budgets to smaller sets of promising candidates. The second, \textbf{AdaEva-R}, builds on statistical \emph{\underline{r}acing}~\citep{birattari2002racing}, which eliminates candidates once sufficient evidence indicates that they are unlikely to outperform their competitors~\citep{lopez2016irace}. Both approaches are particularly well suited to LLM4AD because they do not require predictive models of candidate performance and because they operate directly on instance-level evaluation outcomes.

We empirically demonstrate that AdaEva can substantially improve evaluation efficiency while preserving the effectiveness of the underlying LLM4AD search. Across combinatorial optimization and continuous black-box optimization, we assess search progress throughout the evaluation budget, the quality of the final returned algorithms, and their performance on held-out instances. We further examine how adaptive allocation reshapes evaluation consumption and candidate survival, characterize the resulting evaluation dynamics, and assess the robustness of AdaEva across different LLM backbones, allocation strategies, statistical tests, and racing configurations. Together, these experiments evaluate not only whether adaptive evaluation improves search efficiency, but also how its allocation decisions influence the behavior and outcomes of evolutionary LLM4AD.

Our main contributions can be summarized as follows:
\begin{itemize}

\item We formulate instance-wise candidate evaluation in evolutionary LLM4AD as an adaptive resource-allocation problem and introduce \textbf{AdaEva}, a drop-in evaluation framework that adaptively allocates evaluation effort while preserving the underlying candidate-generation and evolutionary search mechanisms.

\item We instantiate AdaEva with two complementary allocation mechanisms: \textbf{AdaEva-S}, based on successive halving, and \textbf{AdaEva-R}, based on statistical racing. 

\item We show that adaptive evaluation provides more consistent search efficiency and held-out performance under matched evaluation budgets than any single fixed number of evaluations per candidate.

\end{itemize}

\begin{figure}[t]
  \centering
    \includegraphics[width=0.7\textwidth, trim=20 10 20 10, clip]{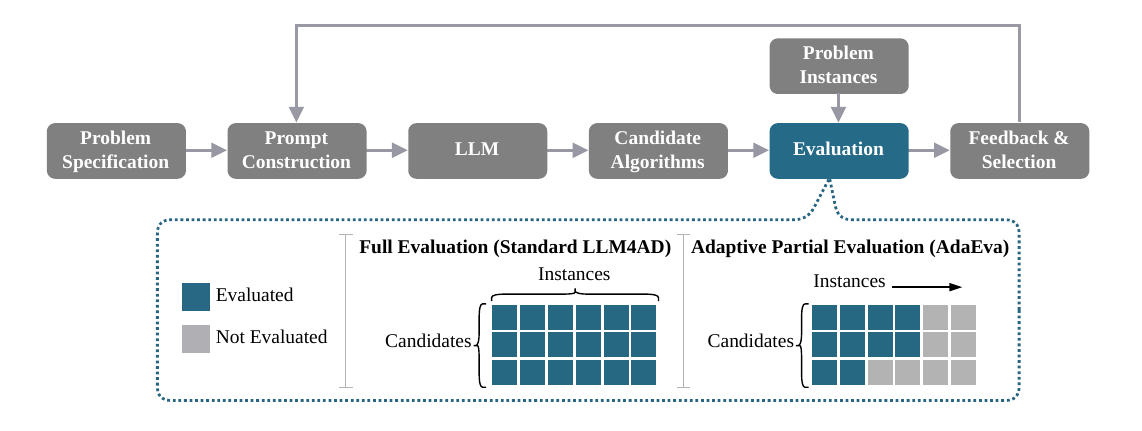}
    \label{fig:AdaEva}
  \caption{LLM-driven algorithm design as a loop. At each generation, a prompt built from the problem specification and selected parents is passed to an LLM, which generates candidate algorithms; these are evaluated on problem instances, and the best are selected and fed back to guide the next generation. Standard frameworks evaluate every candidate on every instance (left). AdaEva replaces only the evaluation stage with adaptive partial evaluation (right): candidates are evaluated progressively, those that fall behind are eliminated early, and the saved budget is concentrated on the candidates that remain competitive. Prompt construction, generation, and selection are unchanged.}
  \label{fig:AdaEva}
\end{figure}

\section{Related Work}
\label{sec:background}

\textbf{LLM-Driven Algorithm Design.}
LLM4AD frameworks use LLMs to generate and revise executable algorithms, with evaluation results guiding subsequent search~\citep{liu2026systematic}. EoH~\citep{liu2024eoh} evolves a population of heuristics through LLM-generated variation and recombination, while ReEvo~\citep{ye2024reevo} uses reflective feedback and memory to guide later generations. LLaMEA~\citep{van2025llamea} applies LLM-guided evolutionary search to the design of black-box optimization algorithms. Subsequent work has explored other ways to guide and organize search: HSEvo~\citep{dat2025hsevo} uses harmony search to balance exploration and exploitation, and MCTS-AHD~\citep{zheng2025monte} structures exploration as a search tree. More recently, HiFo-Prompt~\citep{zhong2026hifo} combines foresight prompts that respond to population dynamics with hindsight prompts that distill reusable insights from earlier generations. Work on generalization has also expanded to more diverse training instances. EoH-S searches for a portfolio of complementary heuristics~\citep{liu2026eoh} on a large set of instances with different distributions and sizes, and MoH~\citep{shi2026generalizable} develops heuristic optimizers through self-invocation and multi-instance training. These advances change how candidates are generated, selected, or retained, but they do not address the question of how to efficiently compare generated candidate algorithms.

\textbf{Evaluation Efficiency in LLM4AD.}
Candidate evaluation is often a major computational bottleneck in LLM-driven search, particularly when each candidate must be executed across many problem instances, stochastic runs, or expensive simulations. This imbalance can be substantial when generation is cheap relative to evaluation: OrderPlace~\citep{mo2026order}, for example, reports that most optimization time is spent evaluating generated strategies, while GEPA~\citep{agrawal2026gepa} similarly highlights subsampled validation as a promising direction for improved efficiency. Existing methods reduce this cost through pre-specified evaluation cascades (AlphaEvolve~\citep{novikov2025alphaevolve}), smaller evaluation subsets (HeuriGym~\citep{chen2026heurigym}), learned fitness predictors or proxy evaluators (Hercules-P~\citep{wu2025efficient}; Janus~\citep{liu2026janus}), or more sample-efficient search policies (ShinkaEvolve~\citep{lange2026shinkaevolve}; AdaEvolve~\citep{cemri2026adaevolve}). In contrast, AdaEva leaves both candidate generation and the per-instance evaluator unchanged, and adaptively determines how much of a shared instance pool each candidate should be evaluated on before allocating further evaluation effort or making a selection decision.

\textbf{Adaptive Partial Evaluation in Adjacent Domains.}
Adaptive allocation of evaluation effort has a long history in automated algorithm configuration and hyper-parameter optimization. Racing methods such as F-Race~\citep{birattari2002racing} and iterated racing~\citep{lopez2016irace} incrementally evaluate candidates on shared instances and eliminate those with sufficient evidence of inferior performance, while successive halving and Hyperband~\citep{jamieson2016non,li2018hyperband} progressively allocate larger budgets to smaller sets of promising configurations. Extensions such as BOHB~\citep{falkner2018bohb} and DEHB~\citep{awad2021dehb} combine such resource allocation with model-based and evolutionary search, and more recent work explicitly treats evaluation as a sequential budget-allocation problem under ranking uncertainty~\citep{zouhar2026dynamically}. AdaEva builds on these principles but uses them as \emph{instance-wise adaptive partial evaluation} mechanisms within existing evolutionary LLM4AD frameworks, leaving their candidate-generation and population-update logic unchanged and thereby decoupling evaluation allocation from the underlying search strategy.

\section{Instance-wise Adaptive Partial Evaluation for LLM4AD} 

\subsection{Evolutionary Loop in LLM4AD}
\label{sec:prelim}
For a given problem $\mathcal{P}$ and a (typically not further specified) space $\mathcal{H}$ of executable algorithms, 
LLM-driven algorithm design casts the search as an evolutionary loop in which an LLM acts as the variation operators (\Cref{fig:AdaEva}). At generation $t$, a population $\Pi^{(t)}\subseteq\mathcal{H}$ of size $M$ supplies parents for a generator $\mathcal{L}$. These parents, together with the problem specification, are used to construct prompts from which the generator produces an offspring set $\mathcal{O}^{(t)}\subseteq\mathcal{H}$.

The current population and offspring are then combined into a candidate pool $\mathcal{U}^{(t)} = \Pi^{(t)} \cup \mathcal{O}^{(t)}$, and the $M$ candidates with the best empirical performance are retained as the elites and copied to the next population.
For this comparison, the performance of an algorithm $h \in \mathcal{H}$ is evaluated as the average quality $\hat{g}(h) =\sum_{x\in\mathcal{S}} f(h;x) / {|\mathcal{S}|}$ that $h$ achieves on a subset $\mathcal{S} \subseteq \mathcal{I}$ of relevant problem instances $\mathcal{I}$, with $f(h;x)$ being the (average, in case of randomized algorithms) solution quality obtained by this heuristic on instance $x \in \mathcal{S}$. 
In standard LLM4AD frameworks, the evaluation is conducted on the whole instance set ($\mathcal{S} \equiv \mathcal{I}$). Note that at each generation, only the offprings need to be evaluated, as their parents' performance were obtained from the previous generation. 

\subsection{Adaptive Evaluation-Budget Allocation}
\label{sec:AdaEva_method}

Evaluating every candidate on the full pool spends the budget uniformly regardless of how candidates perform. Clearly inferior candidates consume as much as competitive ones, while closely matched candidates may need more evidence than either receives. Our proposed AdaEva framework acts as an evaluation layer that instead allocates budget progressively and leaves candidate generation, parent selection, and population update of the underlying LLM4AD framework untouched. We adapt two commonly-used techniques from the adjacent areas of hyper-parameter optimization and automated algorithm configuration. \textbf{AdaEva-S} follows \emph{successive halving}~\citep{karnin2013almost,li2018hyperband} and removes a fixed fraction of candidates after a predetermined number of evaluations, whereas \textbf{AdaEva-R} follows iterated racing~\citep{birattari2002racing,lopez2016irace} and removes candidates using statistical tests whenever the accumulated evidence establishes that they are inferior.

\begin{algorithm}[t]
\small
\caption{
AdaEva for one generation.
AdaEva-S and AdaEva-R differ only in Lines~\ref{line:new_eval} and \ref{line:remove_cands}.
}
\label{alg:adaeva}
\begin{algorithmic}[1]
\Require Candidate set $\mathcal{U}^{(t)}$, evaluation history $\mathcal{E}$,
population size $M$
\Ensure Survivor set $\mathcal{V}^{(t)}$, incumbent $\hat{h}^{(t)}$,
updated history $\mathcal{E}$

\State $\mathcal{V}^{(t)} \gets \mathcal{U}^{(t)}$

\While{$|\mathcal{V}^{(t)}| > M$}
    \State Evaluate candidates in $\mathcal{V}^{(t)}$ on additional instances,
    reusing evaluations stored in $\mathcal{E}$
    \label{line:new_eval}
    \State Update $\mathcal{E}$ with the new evaluation results

    \State Identify inferior candidates
    $\mathcal{X} \subseteq \mathcal{V}^{(t)}$
    according to the elimination rule
    \label{line:remove_cands}

    \State $\mathcal{X} \gets$ at most
    $|\mathcal{V}^{(t)}|-M$ candidates from $\mathcal{X}$
    \State $\mathcal{V}^{(t)} \gets
    \mathcal{V}^{(t)} \setminus \mathcal{X}$
\EndWhile

\State $\hat{h}^{(t)} \gets$ best performing candidate in $\mathcal{V}^{(t)}$ 
\State \Return $\mathcal{V}^{(t)}, \hat{h}^{(t)}, \mathcal{E}$
\end{algorithmic}
\end{algorithm}

\Cref{alg:adaeva} summarizes the common procedure that is used by both AdaEva variants at each generation $t$ of their evolutionary loop. As described in~\Cref{sec:prelim}, starting from the combined candidate set $\mathcal{U}^{(t)}$ of parents and offsprings, the goal is to select $M$ candidates to form the next population $\mathcal{U}^{(t+1)}$. AdaEva performs this selection iteratively. At each round, the current survivor set $\mathcal{V}^{(t)}$ is evaluated on additional problem instances, while reusing all previously collected results stored in the evaluation history $\mathcal{E}$ (line 3). Based on the accumulated evidence, a subset of inferior candidates is then eliminated (line 5), and the procedure repeats on the remaining candidates. Elimination is capped to retain at least $M$ candidates (lines 6--7). When $M$ remain, they form the next population and the best survivor is returned as the incumbent. AdaEva-S and AdaEva-R share this overall structure and differ only in how additional evaluations are allocated and how candidates are eliminated.

Partial evaluation also affects how performance feedback is presented to the LLM. In our main experiments, we retain each framework's native prompting, without indicating that feedback may be based on a subset of instances, thereby isolating the evaluation layer. A partial-evaluation-aware variant shows no consistent advantage (\Cref{app:partial_eval_feedback,app:sensitivity}).


\textbf{AdaEva-S as a scheduled elimination strategy.}
We define the \emph{evaluation depth} of a candidate as the number of problem instances on which it has been evaluated. AdaEva-S follows the resource-allocation principle of successive halving~\citep{karnin2013almost}: candidates are evaluated at progressively larger depths, while a fixed fraction of inferior candidates is eliminated at each round. AdaEva-S has two hyper-parameters: the initial evaluation depth $\underline{n}_0$ and the reduction factor $\eta>1$.
At each evaluation round, candidates are ranked according to their accumulated evaluation results and approximately a $1/\eta$ fraction is retained, subject to keeping at least $M$ candidates. Starting from $\mathcal{U}^{(t)}$, the number of rounds is $J=\left\lceil \log_{\eta}\!\left(|\mathcal{U}^{(t)}|/M\right)\right\rceil$. The evaluation depth increases geometrically across these rounds, from $\underline{n}_0$ instances in the first round to the full instance set $\mathcal{I}$ in the final round. Following the geometrically increasing resource levels commonly used in Hyperband and successive-halving-based multi-fidelity optimization~\citep{li2018hyperband,falkner2018bohb}, we define the cumulative evaluation depth at round $j$, $j=1,\ldots,J$ as 
$ q_j =
\underline{n}_0
\left(
|\mathcal{I}|/\underline{n}_0
\right)^{\frac{j-1}{J-1}}$.  
This way, early rounds use inexpensive partial evaluations to eliminate weak candidates, while progressively more evaluation effort is allocated to those that survive.

\textbf{AdaEva-R as an evidence-driven elimination strategy.}
AdaEva-R follows the principle of iterated racing~\citep{lopez2016irace}, where elimination is driven by accumulated statistical evidence rather than a pre-specified schedule. At each round, every surviving candidate is evaluated on an additional $\Delta$ instances. To account for stochasticity in the generated algorithms, the race may continue after all instances in $\mathcal{I}$ have been evaluated by collecting additional independent evaluations on previously seen instances. Thus, unlike AdaEva-S, reaching evaluation depth $|\mathcal{I}|$ does not necessarily terminate the race. Elimination begins only after candidates reach a minimum evaluation depth $\underline{n}_0$, after which a statistical decision rule is applied at every round. If the observed performance differences are not statistically distinguishable from instance-to-instance and stochastic variation, all candidates remain in the race; otherwise, candidates with sufficient evidence of inferior performance are removed. Consequently, both the evaluation depth at which elimination occurs and the number of candidates eliminated in each round are adaptive and may vary across generations. Accordingly, AdaEva-R also modifies the stopping criterion in~\Cref{alg:adaeva}: besides terminating when $M$ candidates remain, the race stops when no candidate has been eliminated for $\ell$ consecutive rounds.
Alternatively, as in original iterated-racing~\cite{lopez2016irace}, the surviving set can be carried forward without being refilled to size $M$; we study this choice in~\Cref{sec:ablation}.
By default, we follow the racing procedure of~\citet{birattari2002racing}, using the rank-based Friedman test to detect overall performance differences, followed by Conover's post-hoc test to identify inferior candidates at confidence level $\alpha$. We ablate alternative statistical procedures, including a paired $t$-test and the Nemenyi post-hoc test in~\Cref{sec:ablation} and ~\Cref{app:sensitivity}.

\section{Experiments}
\label{sec:experiments}

\begin{table}[t]
\vspace{-10pt}
\centering
\caption{Benchmark and search settings including the search and held-out instance sets, population size $M$, offspring generated per generation $N$, number of generations $T$, and evaluation budget $B$.}
\label{tab:app_benchmark_setup}
\vspace{0.5em}
\small
\setlength{\tabcolsep}{4pt}
\begin{tabular}{ll>{\raggedright\arraybackslash}p{4cm}>{\raggedright\arraybackslash}p{5cm}cccr}
\toprule
\textbf{Framework} & \textbf{Problem} & \textbf{Search Pool} & \textbf{Held-out Set} & $M$ & $N$ & $T$ & $B$ \\
\midrule
\multirow{3}{*}{EoH}
  & OBP
  & 5,000 items, capacity 100
  & 1k/5k/10k items, capacities 100/500
  & 20 & 20 & 20 & 10,000 \\
  & TSP
  & 100-city, uniform coordinates
  & 29 TSPLIB instances
  & 10 & 10 & 10 & 6,400 \\
  & FSSP
  & 50 jobs, 2--20 machines
  & Taillard benchmark
  & 10 & 10 & 10 & 6,400 \\
\midrule
\multirow{3}{*}{HiFo-Prompt}
  & OBP
  & 5,000 items, capacity 100
  & 1k/5k/10k items, capacities 100/500
  & 10 & 50 & 10 & 12,500 \\
  & TSP
  & 100-city, uniform coordinates
  & 29 TSPLIB instances
  & 10 & 50 & 5 & 16,000 \\
  & FSSP
  & 50 jobs, 2--20 machines
  & Taillard benchmark
  & 10 & 50 & 5 & 16,000 \\
\midrule
LLaMEA
  & BBOB
  & 24 func. $\times$ 3 instances, 5 dims
  & 24 func. $\times$ 5 instances, 5/10/20 dims
  & 10 & 10 & 20 & 14,400 \\
\bottomrule
\end{tabular}
\end{table}

\textbf{LLM4AD Frameworks.} We evaluate AdaEva as an off-the-shelf evaluation layer in three representative LLM4AD frameworks: Evolution of Heuristics (EoH)~\citep{liu2024eoh}, HiFo-Prompt~\citep{zhong2026hifo}, and LLaMEA~\citep{van2025llamea}. They differ in how candidates are generated and selected, with EoH applying multiple variation strategies and selecting once per generation, HiFo-Prompt applying its operators sequentially with selection after each, and LLaMEA evolving parent to offspring with a parent's measured performance written into the prompt. They also differ in what is evolved, such as a problem-specific decision rule embedded in a fixed solver for EoH and HiFo-Prompt, and a complete executable optimization algorithm for LLaMEA. Adaptive evaluation is therefore tested across both component-level heuristic design and full-algorithm synthesis. We retain each framework's native generation, variation, population-management, and prompting mechanisms, changing only the evaluation stage. Framework descriptions and their instantiations are in~\Cref{app:frameworks}.

\textbf{Benchmark Settings.} We examine AdaEva across two broad optimization domains, combinatorial optimization (CO) and continuous black-box optimization (BBO). The CO suite comprises Online Bin Packing (OBP)~\citep{seiden2002online}, where items arriving sequentially must be assigned to bins; the Traveling Salesperson Problem (TSP)~\citep{matai2010traveling}, where the goal is a short tour visiting all cities; and Flow Shop Scheduling (FSSP)~\citep{emmons2012flow}, where jobs must be scheduled across multiple machines. The component evolved in each is the bin-scoring function for OBP and the landscape-update heuristic within Guided Local Search (GLS)~\citep{voudouris1997guided} for TSP and FSSP, whereas LLaMEA evolves complete metaheuristic procedures tested on BBOB suite~\citep{hansen2009real}.

\emph{Generalization} remains a central challenge in machine learning and is equally important in LLM-driven algorithm design. We therefore evaluate incumbents recorded at five checkpoints corresponding to 20\%, 40\%, 60\%, 80\%, and 100\% of the total search budget on held-out instances that are disjoint from the search pools and introduce controlled changes in problem scale, instance source, or dimensionality. For OBP, the held-out instances increase problem scale and vary capacity, using $1$k, $5$k, and $10$k items with $C\in\{100,500\}$, which are originally tested in~\citep{romera2024mathematical,liu2024eoh}. For TSP, the search uses synthetic 100-city instances with coordinates sampled from $[0,1]^2$, whereas held-out evaluation uses TSPLIB~\citep{reinelt1991tsplib} instances, introducing a shift in instance source and distribution. For FSSP, the search uses synthetic instances with uniformly sampled processing times, while held-out evaluation uses the external Taillard benchmark~\citep{taillard1993benchmarks}, again changing the instance distribution. For BBOB, the search uses three instances of each function in $d=5$, while held-out evaluation uses disjoint instances in $d\in\{5,10,20\}$; the higher-dimensional settings additionally require the generated algorithm to transfer to larger search spaces and different optimization landscapes. Brief descriptions of the benchmarks are provided in~\Cref{tab:app_benchmark_setup}, with full details given in~\Cref{app:benchmark_setup}.

\textbf{LLM Backbones.} The principal experiments use Devstral~\citep{rastogi2025devstral} as the primary backbone, selected for its open availability and specialization in coding and software-engineering instances. We additionally use the instruction-tuned Qwen2.5-Coder~\citep{hui2024qwen2}, another openly available coding model with strong code-generation, reasoning, and code-fixing capabilities, to test whether the observed effects persist across independently developed LLM generators. This focus on open coding models, however, leaves open whether the findings extend to proprietary frontier models. We therefore evaluate selected settings with Gemini-3.1-Flash-Lite~\citep{team2023gemini}, a proprietary model that supports instruction following and configurable internal thinking, providing a complementary test beyond open-weight coding backbones. All models use a maximum generation length of $2{,}048$ tokens.

\textbf{Baselines.} We compare AdaEva-S and AdaEva-R with Fixed-$K$ evaluation, where each candidate is evaluated on $K\in \{1,|\mathcal{I}|/4,|\mathcal{I}|/2,|\mathcal{I}|\}$ instances. This spans evaluation depths from a single instance to the full pool and tests whether one fixed $K$ is robust across framework--problem settings. Within each benchmark, all strategies use the same evaluation budget, instance pool, initial population, and LLM4AD configuration, repeated with five random seeds.

\textbf{Experimental Configuration.} 
AdaEva-R uses the parameters recommended for \textsc{irace}~\citep{lopez2016irace}: the first test is applied once the evaluation depth reaches $\underline{n}_0=5$ instances, one instance is added per round ($\Delta=1$), elimination is decided using Friedman test followed by Conover post-hoc test at a significant level $\alpha=0.05$, and a race ends after $\ell=2$ rounds without removal or once $M$ candidates remain. AdaEva-S is configured to match AdaEva-R's setting where comparable, with an initial depth $\underline{n}_0=5$ and a reduction factor $\eta=1.25$. Sensitivity to these parameters is examined in~\Cref{sec:ablation}. 

\begin{table}[t]
\vspace{-10pt}
\centering
\caption{Held-out generalization performance of each evaluation strategy across LLM4AD frameworks, benchmarks, and LLM backbones. Mean ranks (lower is better) are computed across held-out instance group, then averaged across checkpoints. \textbf{Bold} and \underline{underlined} denote the best and second-best methods in each row, respectively. Detailed per-checkpoint and third LLM backbone (Gemini) results are provided in~\Cref{app:more_results}.}
\label{tab:heldout_rank}
\vspace{0.5em}
\small
\setlength{\tabcolsep}{5pt}
\begin{tabular}{lllrrrr>{\columncolor{adaevacolor}}r>{\columncolor{adaevacolor}}r}
\toprule
\textbf{Framework} & \textbf{Benchmark} & \textbf{LLM} & \textbf{Fixed-1} & \textbf{Fixed-25\%} & \textbf{Fixed-50\%} & \textbf{Full} & \textbf{AdaEva-S} & \textbf{AdaEva-R} \\
\midrule
\multirow{6}{*}{EoH} & \multirow{2}{*}{FSSP} & Devstral & 4.967 & 3.167 & 3.767 & 3.933 & \underline{3.133} & \textbf{2.033} \\
 &  & Qwen & 4.767 & 4.033 & \underline{2.667} & 4.267 & 2.867 & \textbf{2.400} \\
\cmidrule(lr){2-9}
 & \multirow{2}{*}{OBP} & Devstral & \underline{2.717} & \textbf{1.817} & 3.067 & 5.083 & 3.667 & 4.650 \\
 &  & Qwen & \textbf{1.983} & 4.300 & 3.567 & 3.867 & \underline{3.400} & 3.883 \\
\cmidrule(lr){2-9}
 & \multirow{2}{*}{TSP} & Devstral & 4.333 & 2.967 & 3.133 & 6.000 & \underline{2.900} & \textbf{1.667} \\
 &  & Qwen & 3.933 & \underline{2.767} & 2.933 & 5.833 & 3.367 & \textbf{2.167} \\
\midrule
\multirow{6}{*}{HiFo-Prompt} & \multirow{2}{*}{FSSP} & Devstral & 4.500 & 3.100 & 3.133 & 5.700 & \underline{2.433} & \textbf{2.133} \\
 &  & Qwen & 5.167 & \textbf{2.333} & 3.700 & 4.100 & \underline{2.600} & 3.100 \\
\cmidrule(lr){2-9}
 & \multirow{2}{*}{OBP} & Devstral & 3.250 & \underline{2.783} & 4.183 & 5.100 & 2.933 & \textbf{2.750} \\
 &  & Qwen & 3.667 & \underline{2.933} & 3.767 & 4.400 & \textbf{2.883} & 3.350 \\
\cmidrule(lr){2-9}
 & \multirow{2}{*}{TSP} & Devstral & 5.000 & 3.067 & 2.933 & 6.000 & \textbf{1.900} & \underline{2.100} \\
 &  & Qwen & 6.000 & 3.067 & 2.733 & 4.300 & \underline{2.600} & \textbf{2.300} \\
\midrule
\multirow{2}{*}{LLaMEA} & \multirow{2}{*}{BBOB} & Devstral & 5.867 & 3.833 & 4.067 & 3.200 & \textbf{1.933} & \underline{2.100} \\
 &  & Qwen & 6.000 & 3.967 & 4.400 & 3.167 & \underline{2.067} & \textbf{1.400} \\
\bottomrule
\end{tabular}
\end{table}

\subsection{Generalization to Unseen Instances}
\label{sec:generalization}

\Cref{tab:heldout_rank} compares the generalization performance of different allocation strategies across all 14 combinations of framework, benchmark, and LLM backbone. For each combination, the incumbents recorded at five budget checkpoints are evaluated on held-out instances drawn from a different distribution, the strategies are ranked within each checkpoint and held-out group, and the ranks are then averaged.
 
The racing variant of AdaEva attains the best mean rank in 8 (out of 14) combinations and ranks among the best two in 13, except for the EoH/OBP setting. No fixed evaluation depth achieves this advantage. Full-pool evaluation, the default in these frameworks, is the worst or second-worst strategy in every combination, except for the LLaMEA/BBOB setting. 
Meanwhile, shallower fixed evaluation depths tend to perform better, though not consistently. Fixed-1 is the best strategy on EoH/OBP with the Qwen2.5-Coder backbone (mean rank of $1.983$) but is almost the worst on the other combinations, while neither Fixed-25\% nor Fixed-50\% ranks better than third in more than half of the combinations. Which fixed depth performs well therefore depends on the framework, the benchmark, and the backbone, none of which is known before the search is run, whereas both AdaEva instantiations use a single configuration throughout.

\begin{figure}[t]
  \centering
    \includegraphics[width=0.85\textwidth, trim=20 7 20 10, clip]{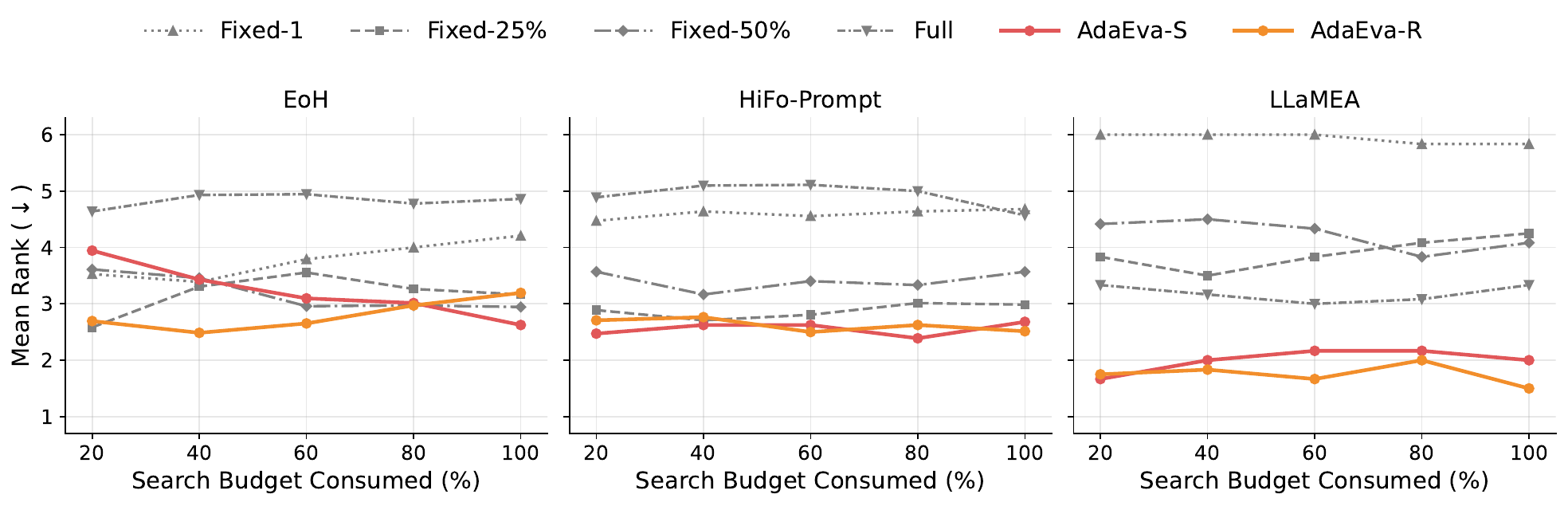}
    \label{fig:anytime_rank_trajectory}

  \caption{Held-out generalization trajectories of evaluation strategies across three frameworks. Mean ranks (lower is better) are computed within each held-out instance group, then averaged across the corresponding benchmarks, and LLM backbones. Each panel shows performance at increasing fractions of the search budget (20\%--100\%).}
  \label{fig:heldout_rank}
\end{figure}

\Cref{fig:heldout_rank} shows that both AdaEva instantiations achieve the strongest held-out ranks across checkpoints. Aggregated over backbones, they rank best at nearly every checkpoint in all three frameworks; on LLaMEA, they lead by more than a full rank from the earliest checkpoint onward. Fixed depths remain stable but lower, while full-pool evaluation improves little as budget accumulates, consistent with completing the fewest generations.

{
\setlength{\OuterFrameSep}{1pt}
\begin{snugshade}
    \textbf{Finding 1.} On unseen instances, AdaEva ranks best or near-best across every framework, benchmark, and LLM backbone tested, whereas no fixed evaluation depth is competitive across them.
\end{snugshade}
}
\vspace{-4pt}

\subsection{Search Progress under Fixed Budget and Adaptive Allocation}
\label{sec:search_progress}

\begin{figure}[t]
  \centering
  \begin{minipage}[b]{0.32\textwidth}
    \includegraphics[width=\textwidth]{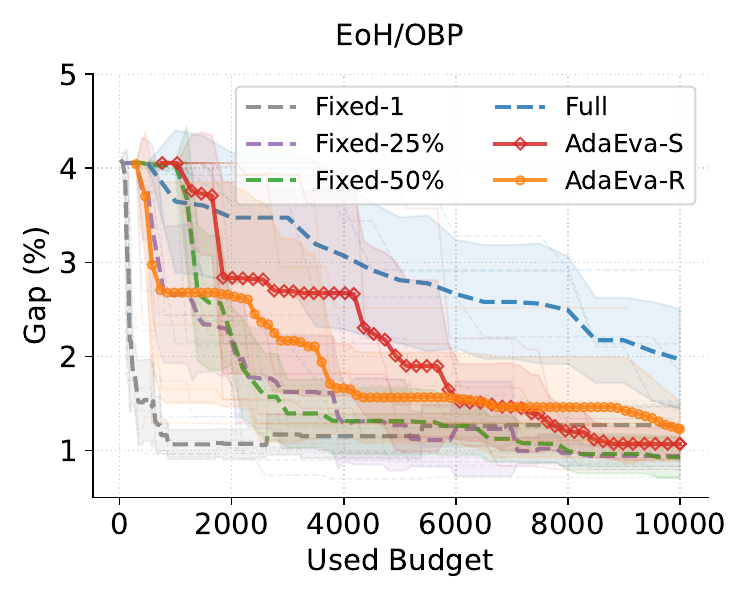}
    \label{fig:eoh_obp_mean_cost_savpop_devstral}
  \end{minipage}
  \begin{minipage}[b]{0.32\textwidth}
    \includegraphics[width=\textwidth]{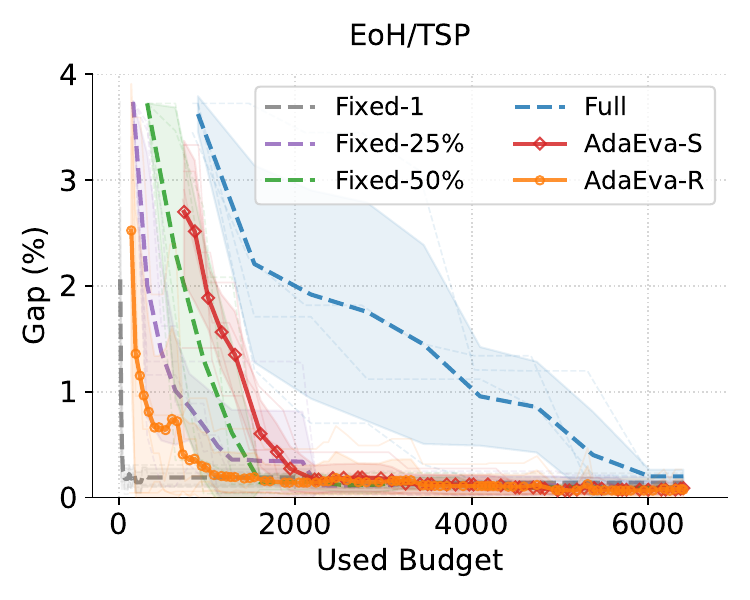}
    \label{fig:eoh_tsp_mean_cost_savpop_devstral}
  \end{minipage}
  \begin{minipage}[b]{0.32\textwidth}
    \includegraphics[width=\textwidth]{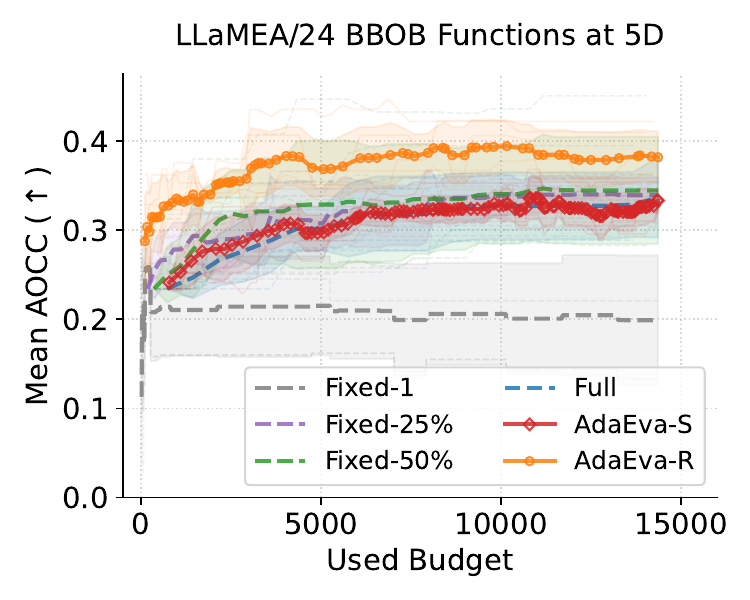}
    \label{fig:llamea_bbob_mean_cost_savpop_pmoriginal_devstral}
  \end{minipage}
\vspace{-15pt}
\caption{Search performance under increasing evaluation budget with the Devstral backbone, on 3 settings chosen to span the range of behaviours observed. Incumbents are selected during search from partial evaluations and then re-evaluated on the complete pool for reporting; the y-axis gives that value, as optimality gap for OBP, gap for TSP, and mean AOCC for BBOB; lower is better in the first two panels and higher is better in the third. Curves show the mean over 5 runs, with shaded regions indicating variability across seeds. The remaining 4 settings, and results with the Qwen2.5-Coder and Gemini-3.1 backbones are available in~\Cref{app:more_results}.}
  \label{fig:valid_curves}
\end{figure}

The held-out results establish which strategies return better algorithms, but not why. We now examine the search itself, reporting the full-pool objective value of the recorded incumbent against budget consumed. A fixed evaluation budget can be spent on many candidates evaluated on few instances each, or on few candidates evaluated on many. We refer to the number of candidates a strategy examines as its \emph{breadth} and the number of instances each receives as its \emph{depth}; their product is fixed by the budget, so a strategy cannot increase one without conceding the other. \Cref{fig:valid_curves} shows the validation curves on three settings chosen to span the behaviours observed, with the remaining four in~\Cref{app:more_results}.

The validation curves show that full-pool evaluation makes the least progress per unit budget in all settings. This is especially clear for EoH/TSP, where it uses nearly the entire budget to reach an objective value attained by other strategies within the first third (\Cref{fig:valid_curves}-middle). Its estimates are the most precise, but this precision shortens the search horizon: each generation costs $640$ evaluations, so few fit within $B=6{,}400$. BBOB (\Cref{fig:valid_curves}-right) is an exception, where full-pool evaluation is mid-ranked because heterogeneous instances make shallow evaluation more prone to misranking candidates.

Reducing the depth helps, but which reduction helps depends on the setting. Fixed-1 is fastest on EoH/OBP and EoH/TSP but by far the weakest on LLaMEA/BBOB, where it reaches a mean AOCC of only $0.20$ versus more than $0.33$ for every other strategy; full-pool evaluation reverses this ordering. The benchmark properties in~\Cref{tab:homo_hetero_analysis} help explain this variation. Kendall's $W$~\citep{kendall1939problem} measures agreement among instance-wise candidate rankings, while the Across- and Within-candidate coefficients of variations~\citep{mckay1932distribution} capture complementary performance variability. On OBP, rankings are identical across instances ($W=1.00$), making shallow evaluation sufficient; TSP is more subtle, with very low ranking agreement ($W=0.03$) but negligible Across CV ($0.0001$) and low Within CV ($0.0348$), consistent with Fixed-1 remaining competitive. BBOB combines weaker ranking agreement ($W=0.66$) with substantially larger Across and Within CVs ($0.68$ and $1.12$), consistent with the poor performance of very shallow evaluation. Thus, no single benchmark statistic determines the useful fixed depth, motivating adaptive evaluation. Further definitions and benchmark-specific analysis of these metrics are provided in~\Cref{app:homo_hetero_analysis}.

Both AdaEva instantiations remain close to the best fixed depth in every panel, and on BBOB AdaEva-R is the strongest strategy overall. They are not uniformly best, and the exceptions are the ones~\Cref{tab:homo_hetero_analysis} predicts. On OBP, where a single instance suffices, Fixed-1 reaches a lower gap than either instantiation; on BBOB, AdaEva-S remains with the mid-ranked group while AdaEva-R leads. What the adaptive strategies avoid is the large loss that each fixed depth incurs in at least one setting, which is what the held-out ranks of~\Cref{sec:generalization} reflect.

\begin{table}[t]
\vspace{-10pt}
\centering
\caption{Benchmark properties and the evaluation depth reached by AdaEva-R. Kendall's $W$ and the Across- and Within-candidate coefficients of variation (CVs) are computed from a fixed initial set of candidates evaluated on every instance. Evaluation depth is reported under EoH for the CO benchmarks and under LLaMEA for BBOB. Definitions of these metrics are provided in~\Cref{app:homo_hetero_analysis}.}
\label{tab:homo_hetero_analysis}
\small
\vspace{0.5em}
\setlength{\tabcolsep}{10pt}
\begin{tabular}{lccccc}
\toprule
& \multicolumn{3}{c}{\textbf{Property}} & \multicolumn{2}{c}{\textbf{AdaEva-R Depth}} \\
\cmidrule(lr){2-4}\cmidrule(lr){5-6}
\textbf{Benchmark} & \textbf{Kendall's $W$} & \textbf{Across CV} & \textbf{Within CV}  & \textbf{Mean Depth} & \textbf{Stop at First Block} \\
\midrule
OBP  & 1.0000 & 0.3403 & 0.0016 & $7.04 \pm 0.72$  & $92.8\% \pm 3.5\%$ \\
TSP  & 0.0312 & 0.0001 & 0.0348 & $12.49 \pm 1.98$ & $73.8\% \pm 5.6\%$ \\
FSSP & 0.6133 & 0.0088 & 0.1216 & $19.25 \pm 1.02$ & $46.1\% \pm 4.4\%$ \\
BBOB & 0.6567 & 0.6809 & 1.1235 & $22.79 \pm 2.94$ & $61.9\% \pm 4.4\%$ \\
\bottomrule
\end{tabular}
\end{table}

\begin{figure}[t]
  \centering
    \includegraphics[width=0.9\textwidth]{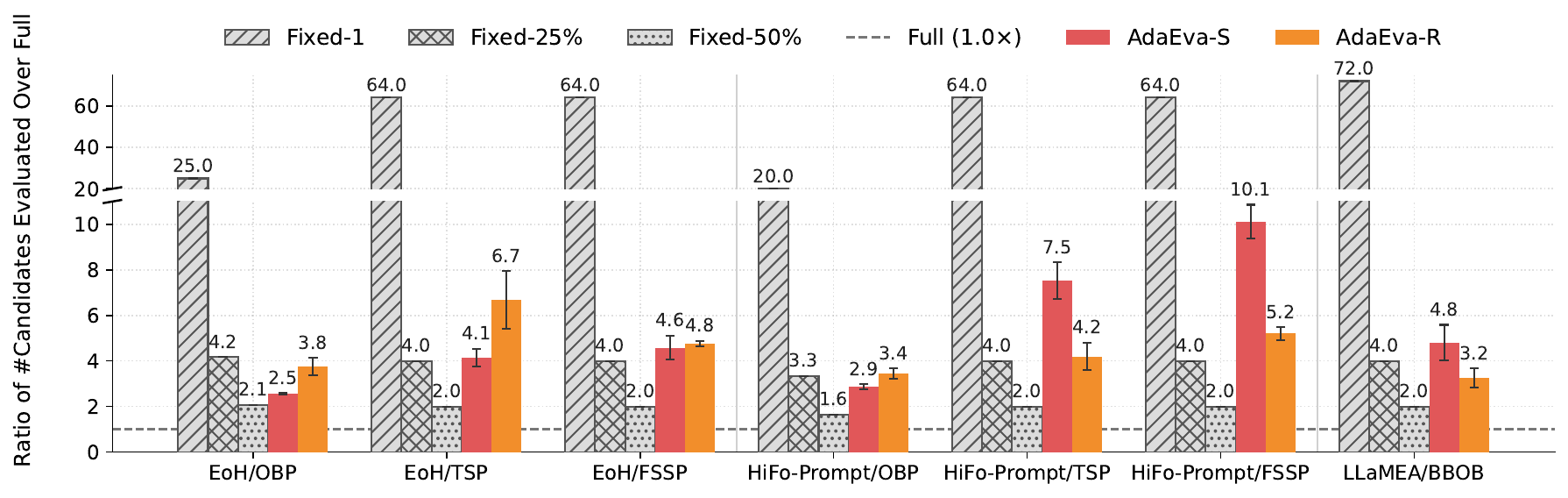}
  \caption{Relative number of candidate algorithms evaluated during the search, under different evaluation strategies across EoH, HiFo-Prompt, and LLaMEA (using the Devstral backbone). The values are normalized by the corresponding full-evaluation baseline, so larger ratios indicate that more candidate algorithms can be sampled within the same evaluation budget; error bars denote variability across five runs.}
  \label{fig:heuristics_ratio_barchart}
\end{figure}

Unlike a fixed depth, which commits to a point on the breadth--depth trade-off before the search begins, AdaEva arrives at one during it. \Cref{fig:heuristics_ratio_barchart} counts the candidates each strategy evaluates, normalized by the number full-pool evaluation affords in the same setting. The fixed strategies sit at ratios determined by their depth alone, $|\mathcal{I}|/K$, identical in every setting and invariant across runs, whereas both AdaEva instantiations fall between the extremes at $2.5$ to $10.1$ times the candidates of full-pool evaluation, differing across benchmarks and between the two instantiations. The depth they reach responds to the benchmark in the direction $W$ implies, as the right columns of~\Cref{tab:homo_hetero_analysis} show that on OBP, where the instances agree, AdaEva-R removes $92.8\%$ of candidates after the first evaluation block at a mean depth of $7.04$ instances, whereas on FSSP and BBOB fewer than two-thirds stop there and the mean depth reaches $19$ and $23$. A fixed depth cannot produce this profile, since it must either pay the depth for every candidate or forgo it for all.

{
\setlength{\OuterFrameSep}{2pt}
\begin{snugshade}
    \textbf{Finding 2.} Full-pool evaluation buys precision at the cost of a shorter search horizon, and the best fixed depth depends on how consistently a benchmark's instances agree about candidates. Adaptive allocation finds a suitable depth on its own, evaluating many candidates shallowly and few deeply.
    
\end{snugshade}
}
\vspace{-4pt}

\subsection{Sensitivity to the AdaEva Configuration}
\label{sec:ablation}

The main experiments use a single default AdaEva configuration across all settings. We assess sensitivity by varying one parameter or design choice at a time, holding the rest fixed, using Devstral on EoH with three CO benchmarks and on LLaMEA/BBOB. Sensitivity is the percent change in incumbent quality relative to the default, computed within matched seeds by taking the median across seeds at each of five budget checkpoints and then averaging across checkpoints; positive values indicate worse performance. \Cref{tab:sensitivity_selected} reports representative configurations, with the full sweep and measurement details in~\Cref{app:sensitivity}.

Most parameter changes have small effects on the three CO benchmarks, although their impact can be substantially larger on BBOB. In particular, EoH/FSSP is relatively insensitive to the tested configurations, with most alternatives improving on or closely matching the default, whereas several BBOB configurations change incumbent quality by more than $20\%$. The AdaEva-S parameters also have different effects across settings: changing $\underline{n}_0$ or $\eta$ has noticeable but modest effects on EoH, while $\eta=1.5$ improves BBOB by $7.74\%$ and $\underline{n}_0=10$ improves it by $2.01\%$. These results suggest that the fixed configuration is generally robust on the EoH benchmarks, while BBOB is more sensitive to the choice of evaluation parameters.
Two changes degrade quality consistently or nearly so. Disabling stochastic evaluation is worse in all four settings, suggesting that drawing fresh seeds once the instance pool is exhausted provides a useful source of additional evaluation evidence. Replacing the Friedman test with the paired $t$-test is worse in three of four settings, with a particularly large degradation on BBOB. Refilling the surviving set with previously eliminated candidates has little effect on the EoH benchmarks but produces a substantial degradation on BBOB when disabled, indicating that its practical importance can depend on the interaction between the race and the host framework. Finally, modifying the LLaMEA prompt to account explicitly for partial evaluation yields a $14.99\%$ degradation on BBOB.

\begin{table}[t]
\vspace{-10pt}
\centering
\small
\caption{Selected sensitivity results. Each value is the percent change in incumbent quality relative to the default of the same instantiation, computed within matched seeds, taking the median across seeds at each of five budget checkpoints and then averaging across checkpoints. Positive values indicate worse performance than the default. The complete sweep is reported in~\Cref{app:sensitivity}.}
\vspace{0.5em}
\label{tab:sensitivity_selected}
\setlength{\tabcolsep}{5pt}
\begin{tabular}{llrrrr}
\toprule
\textbf{Strategy} & \textbf{Configuration} & \textbf{EoH/OBP} & \textbf{EoH/TSP} & \textbf{EoH/FSSP} & \textbf{LLaMEA/BBOB} \\
\midrule
\multirow{2}{*}{AdaEva-S}
& $\underline{n}_0=10$
& $+1.41$ & $-0.17$ & $-0.02$ & $-2.01$ \\
& $\eta=1.5$
& $+0.25$ & $+0.26$ & $-0.02$ & $-7.74$ \\
\midrule
\multirow{4}{*}{AdaEva-R}
& $\mathcal{T}_{\mathrm{stats}}=t$-test
& $+2.39$ & $+0.04$ & $-0.06$ & $+14.93$ \\
& Refill = Off
& $-0.39$ & $+0.03$ & $-0.01$ & $+19.48$ \\
& Stochastic = Off
& $+0.41$ & $+0.09$ & $+0.03$ & $+21.16$ \\
& Prompt = Partial Evaluation
& -- & -- & -- & $+14.99$ \\
\bottomrule
\end{tabular}
\end{table}

\section{Conclusion}

We introduced \textbf{AdaEva}, an adaptive evaluation layer for LLM-driven algorithm design that allocates evaluation effort based on evidence accumulated during search. We study two complementary instantiations: \textbf{AdaEva-S}, based on successive halving~\citep{karnin2013almost,li2017hyperband}, and \textbf{AdaEva-R}, based on statistical racing~\citep{birattari2002racing,lopez2016irace}. Both integrate into existing LLM4AD frameworks without modifying candidate generation, parent selection, or population update. Across three LLM4AD frameworks, multiple LLM backbones, and combinatorial and continuous black-box optimization problems, no single fixed amount of evaluation per candidate performs consistently well across settings. AdaEva instead allocates more evaluation to candidates that remain competitive, balancing search breadth against evaluation effort and yielding more consistent search and held-out performance than any single fixed evaluation strategy.

Our experiments adopt a fixed-budget perspective, asking how well a given evaluation budget is used. An equally natural alternative is fixed-target optimization, which minimizes the evaluation effort required to reach a specified performance level, as common in black-box optimization benchmarking~\cite{HansenABT22}. Our anytime results suggest AdaEva may also be effective in this setting, motivating future evaluation of AdaEva under fixed-target criteria. 

Several questions remain open. First, the relative performance of AdaEva-S and AdaEva-R varies across settings, and our experiments do not identify a reliable criterion for selecting between them in advance. Developing mechanisms that adapt not only the amount of evaluation but also the allocation strategy itself is a promising direction for future work. Second, although the default configurations are relatively stable on the combinatorial benchmarks, some design choices have a larger effect on BBOB, suggesting that broader studies of parameter adaptation may further improve performance. Finally, our experiments isolate candidate evaluation by preserving the native generation, prompting, and population-management mechanisms of each framework and by matching methods according to candidate-evaluation budget. Consequently, we do not separate framework and benchmark effects in a full factorial design, nor do we account for the additional LLM inference cost incurred when cheaper evaluation enables more generations. Extending adaptive resource allocation jointly across evaluation and other components of LLM4AD remains as future work.

\subsection*{Acknowledgments}
This work was supported by the Google Cloud Research Credits program. The project was financially supported by the European Union (ERC, ``dynaBBO'', grant no.~101125586), by ANR project ANR-23-CE23-0035 Opt4DAC, and by an International Emerging Action funded by CNRS Sciences informatiques. This work used the supercomputer at MeSU (\href{https://sacado.sorbonne-universite.fr/plateforme-mesu}{https://sacado.sorbonne-universite.fr/plateforme-mesu}) and Hypatia (\href{https://www.st-andrews.ac.uk/high-performance-computing}{https://www.st-andrews.ac.uk/high-performance-computing}). Tai Nguyen acknowledges funding from the St Andrews Global Doctoral Scholarship programme.

\subsection*{AI use statement}
In this paper, we employed AI to edit the manuscript and produce portions of the code. All AI-assisted outputs were thoroughly reviewed by us. We take responsibility for the final content of this work, including text, claims or artifacts produced with the aid of generative AI.

\bibliography{iclr2027_conference}
\bibliographystyle{iclr2027_conference}

\newpage
\appendix
\begin{center}
\textsc{\Large Supplementary Material for  \\\vspace{.2em} ``AdaEva: Accelerating LLM-Driven Algorithm Design with Adaptive Partial Evaluation''}
\end{center}
\tableofcontents
\addtocontents{toc}{\protect\setcounter{tocdepth}{2}}

\section{LLM4AD Frameworks}
\label{app:frameworks}
This appendix describes how the LLM4AD frameworks used in our experiments instantiate their evolutionary loop. At a high level, each generation consists of parent selection and prompt construction, offspring generation, evaluation, and population update. Parents are drawn from the current population and combined with the problem specification to construct prompts; the generator LLM $\mathcal{L}$ returns offsprings $\mathcal{O}^{(t)}$, which are screened for compilation and interface conformance. Newly generated candidates are evaluated while evaluation results for retained candidates are carried over, and population management determines the candidates retained for subsequent generation. The frameworks differ in how prompts are constructed and how generation and population management are organized; these framework-specific details are described below. AdaEva replaces the evaluation stage while leaving the remaining framework components unchanged.

\paragraph{Evolution of Heuristics (EoH).} EoH~\citep{liu2024eoh} represents each candidate by a natural-language description of its high-level idea, referred to as a \emph{thought}, together with its executable code implementation. The initial population is generated from the problem specification, with one LLM call producing one heuristic. At each subsequent generation, parents are selected from the current population using rank-based sampling, with better-ranked candidates receiving higher selection probability. Five prompt strategies then generate new heuristics: E1 promotes exploration by asking the LLM to produce an algorithm that is substantially different from the selected parents, while E2 asks it to identify the common idea in multiple parents and develop a new heuristic based on that idea. M1 modifies a single parent structurally, M2 changes its parameter settings, and M3 simplifies potentially redundant components. Each strategy is invoked repeatedly, with each LLM call producing one offspring; thus, with population size $M$, the five strategies generate up to $5M$ offspring per generation. These offspring are evaluated and added to the current pool when feasible, after which the best $M$ heuristics are retained for the next generation.

\begin{algorithm}[t]
\caption{One generation of EoH. The evaluation step (Line~\ref{line:eoh_eval}) is the stage AdaEva replaces.}
\label{alg:eoh}
\begin{algorithmic}[1]
\Require Population $\Pi^{(t)}$ of size $M$, prompt strategies $\mathcal{S}=\{\mathrm{E1},\mathrm{E2},\mathrm{M1},\mathrm{M2},\mathrm{M3}\}$, instance pool $\mathcal{I}$
\Ensure Population $\Pi^{(t+1)}$

\State $\mathcal{O}^{(t)} \gets \emptyset$
\ForAll{$s \in \mathcal{S}$ \textbf{in parallel}}
    \For{$j \gets 1$ \textbf{to} $M$}
        \State $\Pi_p \gets \textsc{SelectParents}(\Pi^{(t)})$
        \State $h \gets \mathcal{L}\!\left(\textsc{Prompt}(s,\Pi_p,\mathcal{P})\right)$
        \If{$h$ compiles and satisfies the interface}
            \State $\mathcal{O}^{(t)} \gets \mathcal{O}^{(t)} \cup \{h\}$
        \EndIf
    \EndFor
\EndFor
\State $\mathcal{U}^{(t)} \gets \Pi^{(t)} \cup \mathcal{O}^{(t)}$
\State Evaluate the newly generated candidates in $\mathcal{O}^{(t)}$ on $\mathcal{I}$, retaining prior evaluations for $\Pi^{(t)}$ \label{line:eoh_eval}
\State $\Pi^{(t+1)} \gets \operatorname{Best}_{M}\!\left(\mathcal{U}^{(t)};\hat{g}\right)$
\State \Return $\Pi^{(t+1)}$
\end{algorithmic}
\end{algorithm}
 
\paragraph{Hindsight-Foresight Prompt (HiFo-Prompt).} HiFo-Prompt~\citep{zhong2026hifo} retains EoH's thought-and-code representation and its five foundational evolutionary strategies, while augmenting each prompt with two additional sources of guidance. The \emph{hindsight} component maintains an Insight Pool of design principles distilled from high-performing heuristics; retrieved insights are injected into subsequent prompts, and their utility is updated according to the performance of the offspring they helped generate. The \emph{foresight} component, implemented by the Evolutionary Navigator, monitors recent population progress, stagnation, and description-level diversity, and selects one of three regimes of explore, exploit, or balance, whose corresponding design directive is added to the prompt. Unlike EoH's generation-level application of its five operators, HiFo-Prompt applies its evolutionary operators sequentially. For each operator, the LLM is invoked repeatedly to generate an offspring batch of size $M$, where each offspring is produced by an independent LLM call. These offspring are merged with the current population and immediately subjected to elitist population management, reducing the population back to size $M$ before the next operator is applied. Consequently, survivors of an earlier operator can serve as parents for subsequent operators within the same generation, while each operator operates on a population of size $M$ rather than on the accumulated candidates from preceding operators.

\begin{algorithm}[t]
\caption{One generation of HiFo-Prompt. Selection is applied after each operator rather than once per generation. AdaEva replaces the evaluation step (Line~\ref{line:hifo_eval}); the quantities marked $\dagger$ consume the resulting costs and are therefore affected by partial evaluation, since candidates surviving longer are evaluated on more instances. The generation prompt itself carries no fitness values.}
\label{alg:hifo}
\begin{algorithmic}[1]
\Require Population $\Pi^{(t)}$ of size $M$, operators $\mathcal{S}=\{\mathrm{E1},\mathrm{E2},\mathrm{M1},\mathrm{M2},\mathrm{M3}\}$ with weights $\{w_s\}$, insight pool $\mathcal{K}$, instance pool $\mathcal{I}$
\Ensure Population $\Pi^{(t+1)}$

\State $\Pi \gets \Pi^{(t)}$
\ForAll{$s \in \mathcal{S}$}
    \If{$u \sim \operatorname{Uniform}(0,1) < w_s$}
        \State $\mathcal{O} \gets \emptyset$
        \For{$j \gets 1$ \textbf{to} $M$}
            \State $\Pi_p \gets \textsc{SelectParents}(\Pi;\hat{g})$ \Comment{$\dagger$ rank-based on cost}
            \State $\kappa \gets \textsc{RetrieveInsights}(\mathcal{K})$ \Comment{$\dagger$ utilities scored by cost}
            \State $\theta \gets \textsc{Regime}(\Pi;\hat{g})$ \Comment{$\dagger$ stagnation from best cost}
            \State $h \gets \mathcal{L}\!\left(\textsc{Prompt}(s,\Pi_p,\mathcal{P},\kappa,\theta)\right)$
            \If{$h$ compiles and satisfies the interface}
                \State $\mathcal{O} \gets \mathcal{O} \cup \{h\}$
            \EndIf
        \EndFor
        \State $\mathcal{U} \gets \Pi \cup \mathcal{O}$
        \State Evaluate the newly generated candidates in $\mathcal{O}$ on $\mathcal{I}$, retaining prior evaluations for $\Pi$ \label{line:hifo_eval}
        \State $\Pi \gets \operatorname{Best}_{M}\!\left(\mathcal{U};\hat{g}\right)$
        \State $\mathcal{K} \gets \textsc{UpdateInsights}(\mathcal{K},\mathcal{O},\hat{g})$ \Comment{$\dagger$ credit from cost}
    \EndIf
\EndFor
\State \Return $\Pi^{(t+1)} \gets \Pi$
\end{algorithmic}
\end{algorithm}

\paragraph{Large Language Model Evolutionary Algorithm (LLaMEA).}
LLaMEA~\citep{van2025llamea} generates complete optimization algorithms as executable code and evolves them through an evolution-strategy-like loop. In its original formulation, LLaMEA uses a single-solution population with $\mu=\lambda=1$: one parent algorithm is used to generate one offspring algorithm at each iteration. The offspring is produced by prompting the LLM with the selected parent, its performance information, and a condensed history of previously generated algorithms and scores, and the LLM is instructed either to refine or redesign the algorithm. Compilation or runtime errors are incorporated into the feedback provided to the LLM. Two selection variants are considered: an elitist $(\mu+\lambda)$ strategy and a non-elitist $(\mu,\lambda)$ strategy, both with $\mu=\lambda=1$. In the elitist variant, the best-so-far algorithm is retained as the parent and replaced only when the offspring is at least as good; in the non-elitist variant, the newly generated offspring becomes the parent for the next iteration regardless of its performance. Although the original formulation is single-solution based, the same parent--offspring evolution and selection mechanism naturally extends to population-based evolution by allowing $\mu>1$ and $\lambda>1$.

\begin{algorithm}[t]
\caption{One generation of LLaMEA. The original formulation uses $\mu=\lambda=1$; we use its elitist $(M{+}N)$ mechanism in a population-based extension with $M,N>1$. AdaEva replaces the evaluation step (Line~\ref{line:llamea_eval}); the line marked $\dagger$ writes a parent's measured cost directly into the prompt, so this framework exposes partial estimates to the generator.}
\label{alg:llamea}
\begin{algorithmic}[1]
\Require Population $\Pi^{(t)}$ of size $M$, offspring count $N$, selection mode $\varsigma\in\{(M{+}N),(M,N)\}$, instance pool $\mathcal{I}$, incumbent $\hat{h}$
\Ensure Population $\Pi^{(t+1)}$, updated incumbent $\hat{h}$

\State $\mathcal{O}^{(t)} \gets \emptyset$
\For{$j \gets 1$ \textbf{to} $N$}
    \State $h_p \gets \textsc{SelectParent}(\Pi^{(t)})$
    \State $h \gets \mathcal{L}\!\left(\textsc{Prompt}(h_p,\hat{g}(h_p),\varepsilon_p,\mathcal{P})\right)$ \Comment{$\dagger$ cost in prompt; $\varepsilon_p$: errors}
    \If{$h$ compiles and satisfies the interface}
        \State $\mathcal{O}^{(t)} \gets \mathcal{O}^{(t)} \cup \{h\}$
    \EndIf
\EndFor

\If{$\varsigma=(M{+}N)$}
    \State $\mathcal{U}^{(t)} \gets \Pi^{(t)} \cup \mathcal{O}^{(t)}$ \Comment{elitist: incumbents compete}
\Else
    \State $\mathcal{U}^{(t)} \gets \mathcal{O}^{(t)}$ \Comment{non-elitist: incumbents discarded}
\EndIf

\State Evaluate the newly generated candidates in $\mathcal{O}^{(t)}$ on $\mathcal{I}$, retaining prior evaluations for $\Pi^{(t)}$ \label{line:llamea_eval}
\State $\Pi^{(t+1)} \gets \operatorname{Best}_{M}\!\left(\mathcal{U}^{(t)};\hat{g}\right)$
\State $\hat{h} \gets \operatorname{Best}_{1}\!\left(\{\hat{h}\}\cup\Pi^{(t+1)};\hat{g}\right)$
\State \Return $\Pi^{(t+1)}$, $\hat{h}$
\end{algorithmic}
\end{algorithm}

\section{Benchmark Settings}

\label{app:benchmark_setup}

\paragraph{Combinatorial Optimization.}
EoH and HiFo-Prompt are evaluated on Online Bin Packing (OBP), the Traveling Salesperson Problem (TSP), and Flow Shop Scheduling (FSSP), following the benchmark formulations used by EoH. For OBP, the original EoH search uses five Weibull instances with $5{,}000$ items and capacity $C=100$, with fitness computed from the ratio between the lower bound and the number of bins used. We enlarge this search pool to 25 independently generated instances while retaining the same distribution and problem size. For TSP, the original search pool contains 64 randomly generated 100-city instances with coordinates sampled from $[0,1]^2$, and fitness is the average gap to the Concorde optimum. For FSSP, the original search pool contains 64 randomly generated instances with 50 jobs and 2--20 machines, with processing times sampled uniformly from $[0,1]$; fitness is the average makespan. Candidate execution for TSP and FSSP uses the original GLS-based implementations, with at most $1{,}000$ local-search iterations and a 60-second limit per candidate--instance evaluation.

For generalization, we use held-out evaluation sets that are disjoint from the search instances. For OBP, the original EoH study evaluates five independently generated instances for each combination of problem size $1$k, $5$k, and $10$k and capacity $C\in\{100,500\}$, reporting percentage gap to the lower bound. We use these size and capacity settings for held-out evaluation. For TSP, EoH evaluates additional random instances at 20, 50, and 100 cities and 29 TSPLIB instances; the latter are evaluated by gap to the TSPLIB best-known solutions, and we use TSPLIB for our held-out evaluation. For FSSP, EoH evaluates Taillard instance sets with 10 instances per job--machine configuration and reports relative gap to the Taillard reference values; we use Taillard instances for held-out evaluation.

\paragraph{Continuous Black-Box Optimization.}
LLaMEA is evaluated on the BBOB suite~\citep{hansen2009real}. The search pool contains three instances (IDs in \{1,2,3\}) of each of the 24 noiseless BBOB functions in five dimensions, giving 72 function instances in total. BBOB comprises five function groups: separable, low- or moderately-conditioned, highly-conditioned unimodal, multimodal with adequate global structure, and multimodal with weak global structure. Candidate performance is aggregated using the normalized Area Over the Convergence Curve (AOCC).

For generalization, the original LLaMEA study evaluates the generated algorithms on five BBOB instances per function in dimensions $d\in\{5,10,20\}$. We follow the same dimensions and number of held-out instances per function, but use explicitly disjoint instance identifiers (IDs in \{4,5,6,7,8\}) for our evaluation. Each held-out instance is evaluated 3 times in our study using independent evaluation seeds.

\section{More Results}
\label{app:more_results}

This section reports results omitted from the main text for space. \Cref{app:homo_hetero_analysis} analyzes benchmark heterogeneity using Kendall's $W$ and the Across- and Within-candidate coefficients of variation (CVs), providing additional context for the benchmark-dependent evaluation-depth results discussed in the main text. \Cref{app:more_search} gives search-progress curves for the settings and LLM backbones not shown in~\Cref{fig:valid_curves}. \Cref{app:more_heldout} reports held-out performance at every budget checkpoint, first as mean ranks per framework and then as objective values for each benchmark and instance group. 

\subsection{Benchmark Heterogeneity}
\label{app:homo_hetero_analysis}

\paragraph{Metrics.}
To characterize how consistently benchmark instances distinguish between candidate algorithms, we compute Kendall's coefficient of concordance $W$~\citep{kendall1939problem} on a fixed set of candidates evaluated on every instance. Each instance ranks the candidates by objective value, and $W$ measures agreement among these rankings, with $W=1$ indicating identical rankings across instances and $W=0$ indicating no concordance. Candidates that fail on any instance are excluded so that all instances rank the same candidate set.

We additionally report two coefficients of variation (CVs)~\citep{mckay1932distribution}. Across CV is the coefficient of variation of candidates' mean performance over the instance pool, measuring how differentiated the candidates are on average. Within CV is the mean candidate-level coefficient of variation across instances, measuring how much an individual candidate's performance varies across the pool. These statistics are computed from the same complete candidate--instance matrix and are independent of the evaluation strategy.

\paragraph{Discussion.}
The four benchmarks exhibit substantially different combinations of ranking agreement and performance variability, as summarized in~\Cref{tab:homo_hetero_analysis}. OBP has perfect ranking concordance ($W=1.00$), indicating that the relative ordering of candidates is invariant across the evaluated instances. Its Across CV is nevertheless non-negligible ($0.34$), reflecting meaningful differences in candidates' mean performance. This combination is consistent with the strong performance of Fixed-1, although candidates differ substantially in quality, their relative ordering can be recovered from a single instance.

TSP provides a contrasting case. Its ranking concordance is extremely low ($W=0.03$), but its Across CV is nearly zero ($0.0001$) and its Within CV is also small ($0.0348$). Thus, instances can disagree strongly in their fine-grained candidate rankings even though candidate mean performance is tightly concentrated and individual candidates exhibit relatively little variation across instances. This helps explain why the low $W$ does not translate into a clear need for deeper evaluation, and why Fixed-1 remains competitive on TSP.

FSSP lies between these cases, with moderate ranking concordance ($W=0.61$) and a larger Within CV ($0.12$). Its instances therefore exhibit more variation in candidate performance than TSP, making additional evaluation more informative. This is reflected in the greater benefit of intermediate or deeper fixed evaluation depths.

BBOB combines moderate ranking concordance ($W=0.66$) with substantially larger Across and Within CVs ($0.68$ and $1.12$). Candidate performance therefore varies considerably both across candidates and across instances. In this setting, a single-instance estimate provides substantially less information about overall candidate quality, consistent with the poor performance of Fixed-1 and the advantage of deeper or adaptive evaluation.

Overall, these results show that no single benchmark statistic determines the appropriate evaluation depth. Kendall's $W$ captures the consistency of candidate rankings across instances, whereas the CVs capture the magnitude of performance variability. Their combination provides a more informative characterization of when shallow evaluation is likely to be reliable, but does not prescribe a fixed depth across benchmarks.

\subsection{Search Progress Under Fixed Budget}
\label{app:more_search}

\begin{figure}[t]
  \centering
  \begin{minipage}[b]{0.32\textwidth}
    \includegraphics[width=\textwidth]{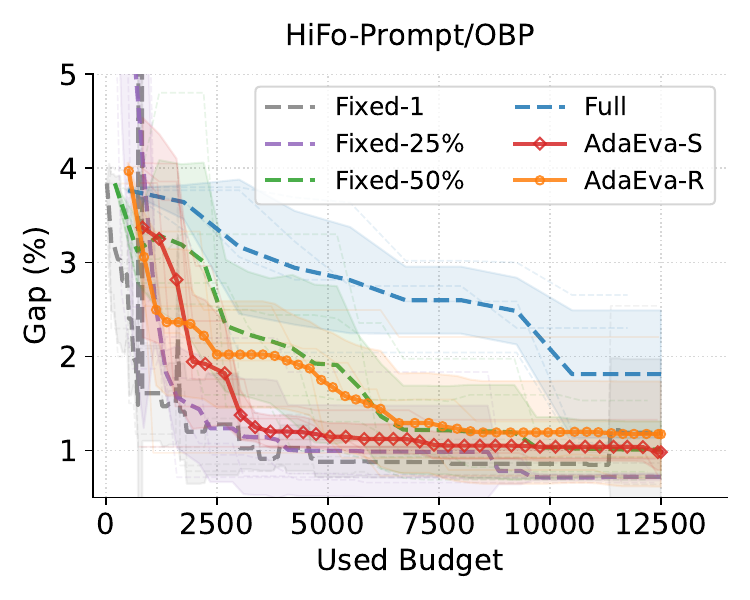}
    \label{fig:hifo_obp_mean_cost_savpop_devstral}
  \end{minipage}
  \begin{minipage}[b]{0.32\textwidth}
    \includegraphics[width=\textwidth]{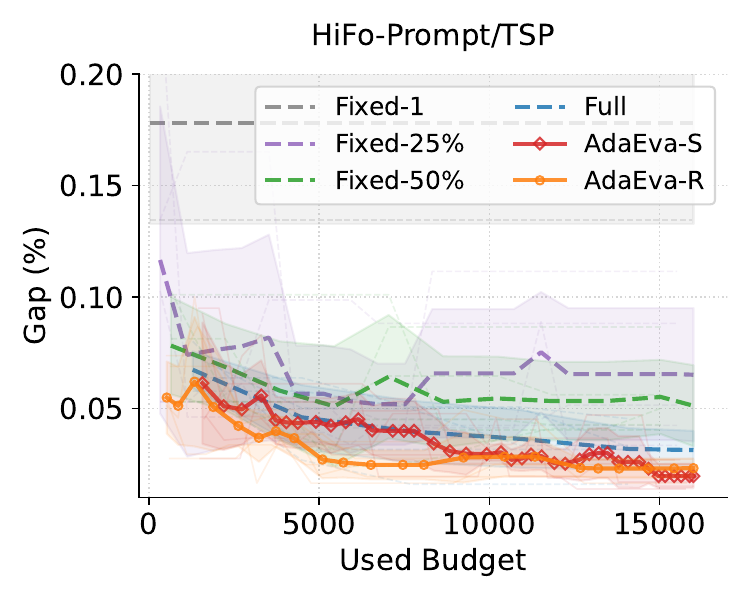}
    \label{fig:hifo_tsp_mean_cost_savpop_devstral}
  \end{minipage}
  \begin{minipage}[b]{0.32\textwidth}
    \includegraphics[width=\textwidth]{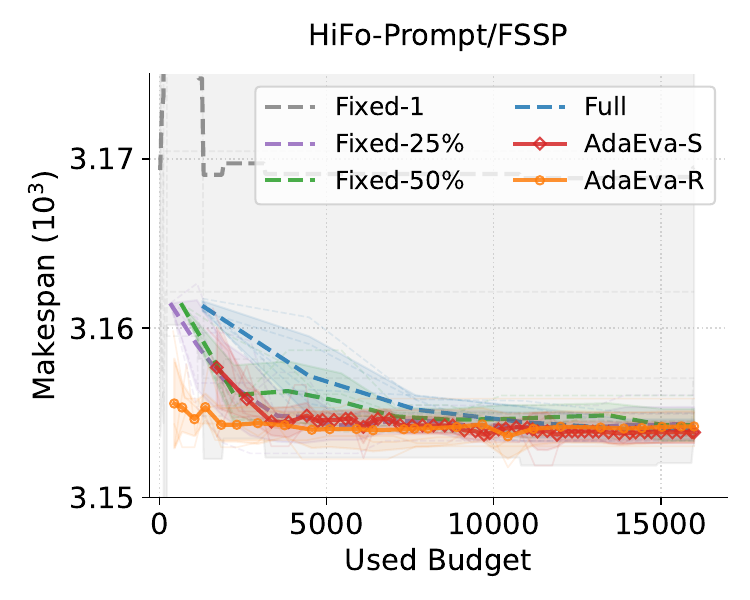}
    \label{fig:hifo_fssp_mean_cost_savpop_devstral}
  \end{minipage}
  \caption{Search performance under increasing evaluation budget with the Devstral backbone for HiFo-Prompt. Each panel compares AdaEva-S and AdaEva-R with Fixed-$K$ evaluation at $K=1$, $25\%$, and $50\%$ of the pool, and with full-pool evaluation ($K=|\mathcal{I}|$). Incumbents are selected during search from partial evaluations and re-evaluated on the complete pool for reporting. The y-axis shows this full-pool performance, measured as optimality gap for OBP and TSP and makespan for FSSP, with lower values better. Curves show mean performance across five independent search seeds, with shaded regions indicating variability across seeds.}
  \label{fig:valid_curves_hifo_devstral}
\end{figure}

\begin{figure}[t]
  \centering
  \begin{minipage}[b]{0.32\textwidth}
    \includegraphics[width=\textwidth]{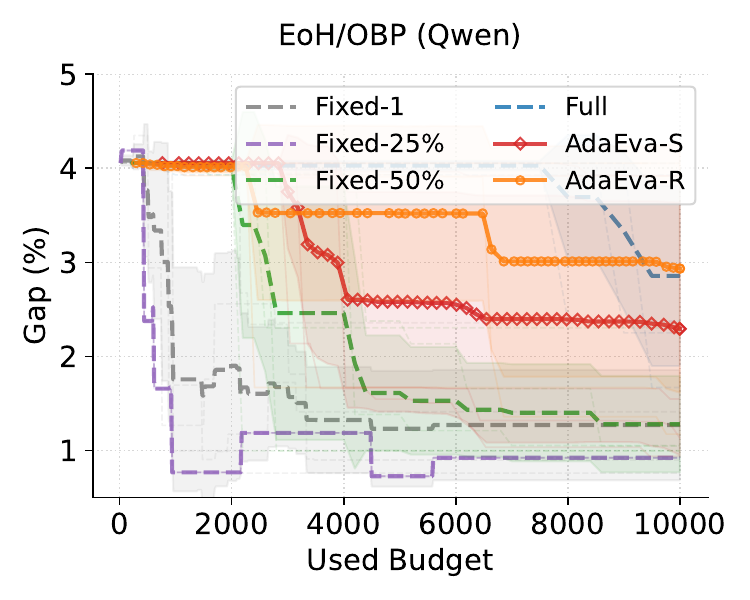}
    \label{fig:eoh_obp_mean_cost_savpop_devstral}
  \end{minipage}
  \begin{minipage}[b]{0.32\textwidth}
    \includegraphics[width=\textwidth]{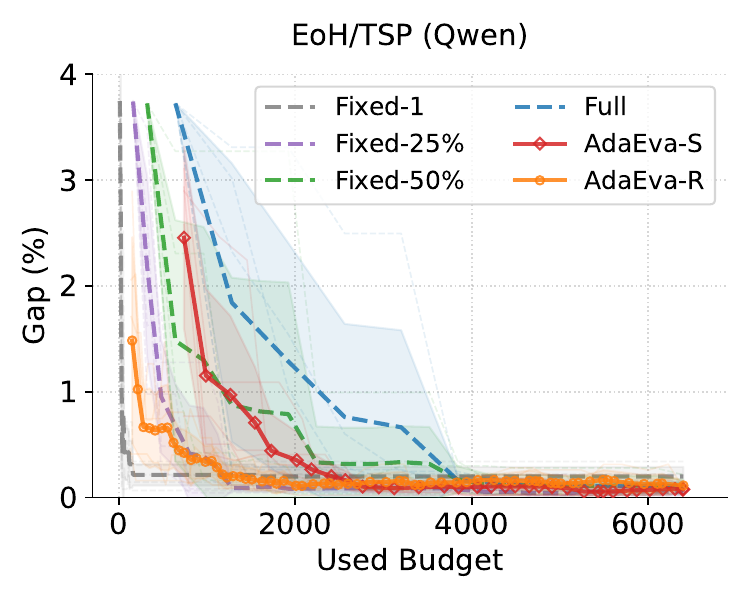}
    \label{fig:eoh_tsp_mean_cost_savpop_devstral}
  \end{minipage}
  \begin{minipage}[b]{0.32\textwidth}
    \includegraphics[width=\textwidth]{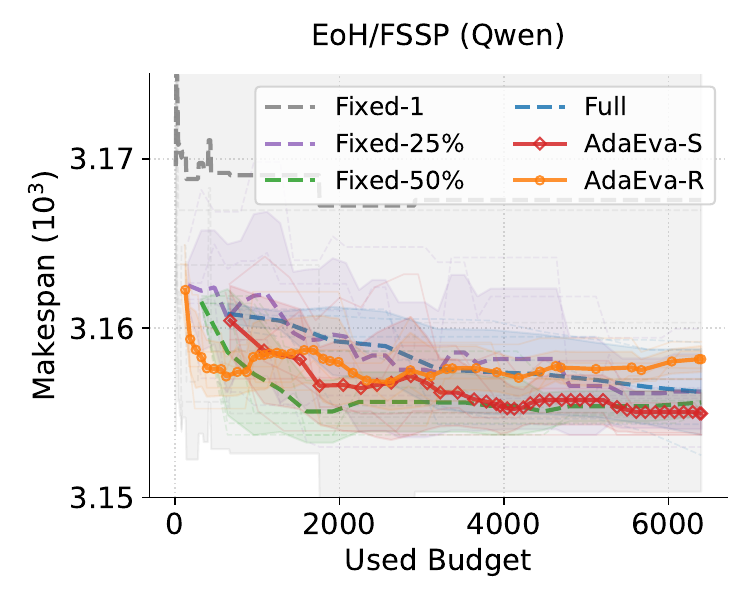}
    \label{fig:eoh_fssp_mean_cost_savpop_devstral}
  \end{minipage}

  \begin{minipage}[b]{0.32\textwidth}
    \includegraphics[width=\textwidth]{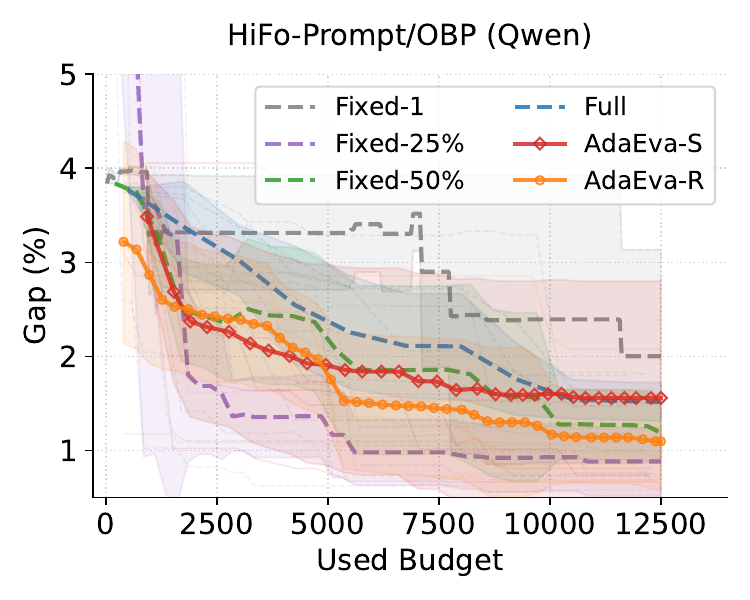}
    \label{fig:hifo_obp_mean_cost_savpop_devstral}
  \end{minipage}
  \hfill
  \begin{minipage}[b]{0.32\textwidth}
    \includegraphics[width=\textwidth]{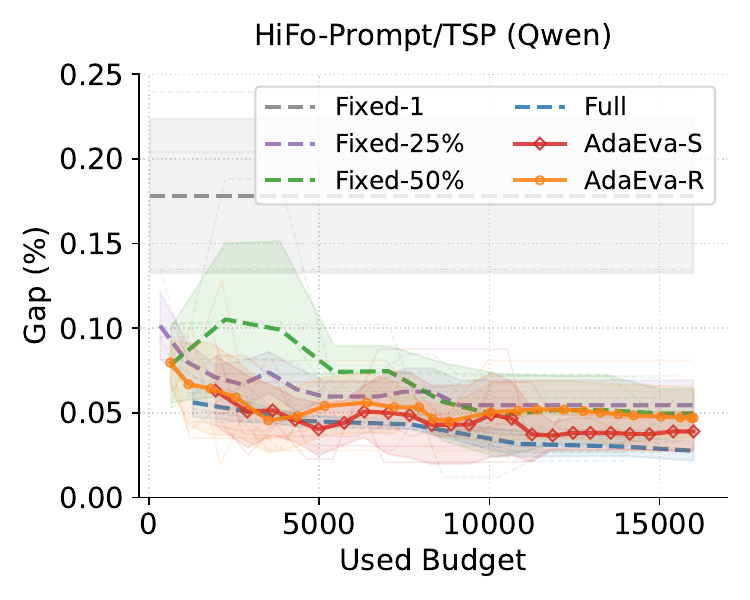}
    \label{fig:hifo_tsp_mean_cost_savpop_devstral}
  \end{minipage}
  \hfill
  \begin{minipage}[b]{0.32\textwidth}
    \includegraphics[width=\textwidth]{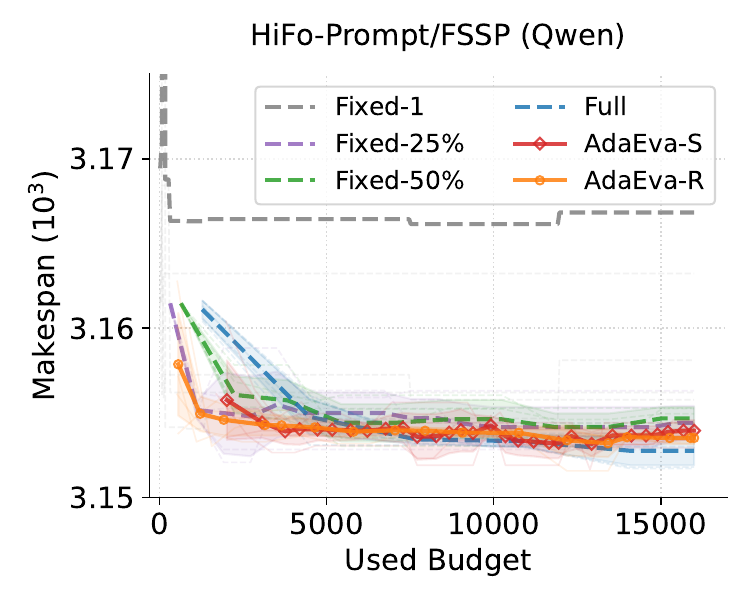}
    \label{fig:hifo_fssp_mean_cost_savpop_devstral}
  \end{minipage}
  \caption{Search performance under increasing evaluation budget with the Qwen2.5-Coder backbone for EoH (top) and HiFo-Prompt (bottom). Each panel compares AdaEva-S and AdaEva-R with Fixed-$K$ evaluation at $K=1$, $25\%$, and $50\%$ of the pool, and with full-pool evaluation ($K=|\mathcal{I}|$). Incumbents are selected during search from partial evaluations and re-evaluated on the complete pool for reporting. The y-axis shows this full-pool performance, measured as optimality gap for OBP and TSP and makespan for FSSP, with lower values better. Curves show mean performance across five independent search seeds, with shaded regions indicating variability across seeds.}
  \label{fig:eoh_hifo_valid_curves_qwen}
\end{figure}

\begin{figure}[t]
  \centering
  \begin{minipage}[b]{0.33\textwidth}
    \includegraphics[width=\textwidth]{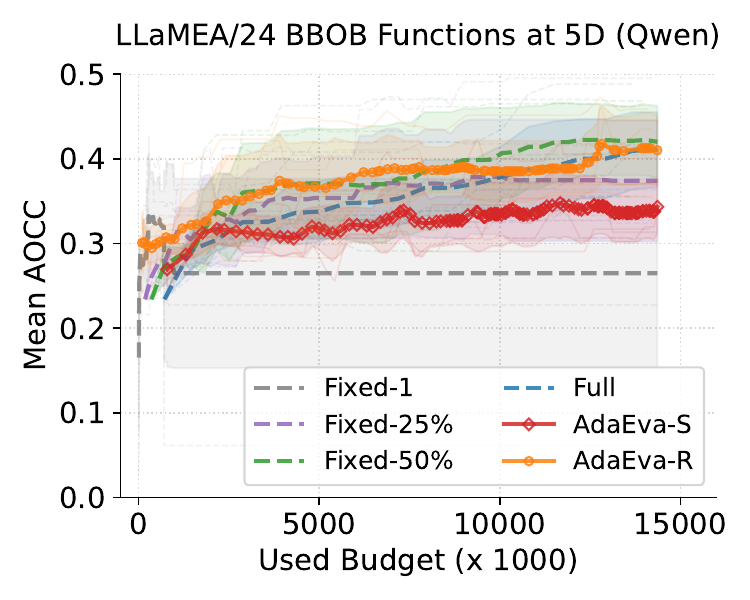}
    \label{fig:llamea_bbob_mean_cost_savpop_pmoriginal_qwen}
  \end{minipage}
  \begin{minipage}[b]{0.33\textwidth}
    \includegraphics[width=\textwidth]{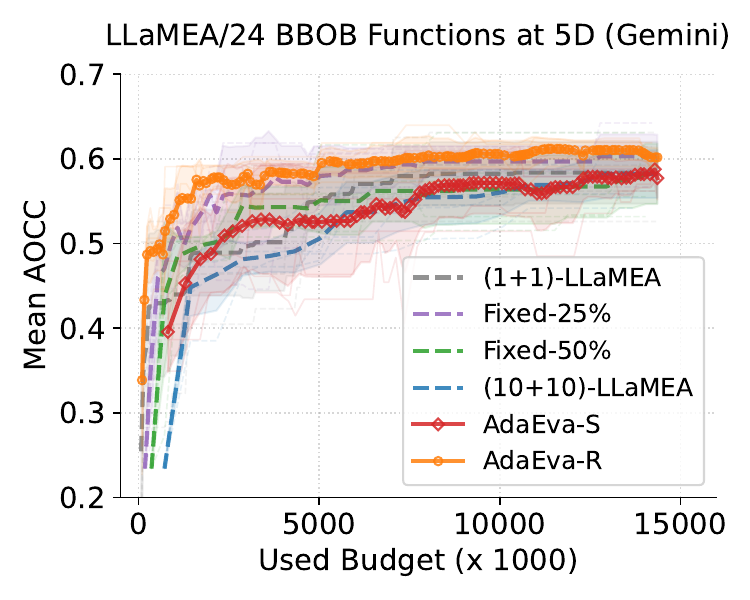}
    \label{fig:llamea_bbob_mean_cost_savpop_pmoriginal_gemini}
  \end{minipage}
  \begin{minipage}[b]{0.33\textwidth}
    \includegraphics[width=\textwidth]{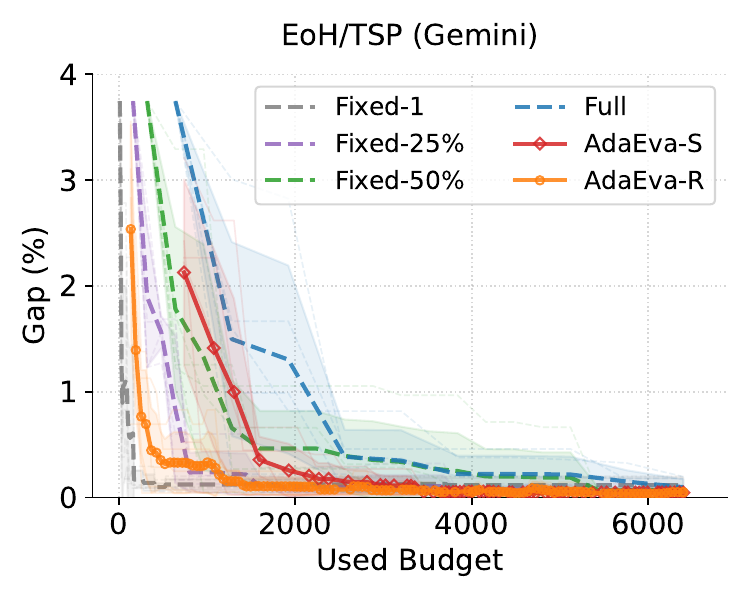}
    \label{fig:eoh_tsp_mean_cost_savpop_gemini}
  \end{minipage}
  \caption{Search performance under increasing evaluation budget with the Qwen2.5-Coder and Gemini-3.1-Flash-Lite backbones. The LLaMEA/BBOB panels and the EoH/TSP panel compare AdaEva-S and AdaEva-R with Fixed-$K$ evaluation at $K=25\%$ and $50\%$ of the pool, and with full-pool evaluation ($K=|\mathcal{I}|$). Incumbents are selected during search from partial evaluations and re-evaluated on the complete pool for reporting. The y-axis shows mean AOCC for BBOB and optimality gap for TSP; higher AOCC and lower gap indicate better performance. Curves show mean performance across five independent search seeds, with shaded regions indicating variability across seeds.}
  \label{fig:llamea_valid_curves_qwen_gemini}
\end{figure}

\Cref{fig:valid_curves} in the main text shows three of the seven framework--benchmark settings with the Devstral backbone. \Cref{fig:valid_curves_hifo_devstral} completes that backbone with the three HiFo-Prompt settings, while~\Cref{fig:eoh_hifo_valid_curves_qwen} repeats the EoH and HiFo-Prompt comparisons with Qwen2.5-Coder. For the additional LLM backbones,~\Cref{fig:llamea_valid_curves_qwen_gemini} shows LLaMEA/BBOB with Qwen2.5-Coder and Gemini-3.1-Flash-Lite, together with EoH/TSP using Gemini-3.1-Flash-Lite.

The additional search-progress curves largely preserve the qualitative pattern observed in the main text. Across the CO settings, full-pool evaluation generally progresses most slowly, while the adaptive methods reach competitive full-pool quality earlier. The Qwen2.5-Coder-based curves do not show a systematic reversal of this behavior relative to Devstral's results, although the absolute performance and the relative separation between strategies vary by benchmark. On LLaMEA/BBOB, both Qwen2.5-Coder and Gemini-3.1 likewise show sustained gains from adaptive evaluation as the budget increases.

\subsection{Generalization to Unseen Instances}
\label{app:more_heldout}

\subsubsection{Mean Ranks per Checkpoint}
\Cref{tab:heldout_rank} in the main text macro-averages held-out ranks over the five budget checkpoints. \Cref{tab:checkpoint_eoh_devstral_qwen,tab:checkpoint_hifo_devstral_qwen,tab:llamea_bbob_main} report these ranks at each checkpoint separately, one table per framework, together with the overall mean rank and the proportions of comparison cells in which each strategy is ranked first or among the top two.

The checkpoint breakdown reveals how the relative performance of the adaptive methods evolves during search. Under EoH (\Cref{tab:checkpoint_eoh_devstral_qwen}), AdaEva-R has the best mean rank at the first three checkpoints and the best overall rank of $2.800$. AdaEva-S improves steadily from $3.944$ at $20\%$ of the budget to the best rank of $2.625$ at $100\%$, overtaking AdaEva-R at the final checkpoint. This difference is consistent with their respective evaluation mechanisms: AdaEva-S follows a predetermined schedule and reaches the full evaluation depth only at its final stage, whereas AdaEva-R allocates further evaluation according to the evidence accumulated during the race.

Under HiFo-Prompt (\Cref{tab:checkpoint_hifo_devstral_qwen}), no such crossover occurs. AdaEva-S and AdaEva-R remain the top two strategies at every checkpoint, with overall ranks of $2.558$ and $2.622$, respectively. Their ranks also remain close throughout the search, indicating that both evaluation mechanisms provide similar held-out performance in this framework.

The LLaMEA results (\Cref{tab:llamea_bbob_main}) show the clearest separation. Both AdaEva instantiations remain below a mean rank of $2.3$ at every checkpoint, with AdaEva-R achieving the best overall rank of $1.850$ and AdaEva-S $2.017$. Together, they account for the best-ranked strategy in $96.7\%$ of comparison cells. In contrast, Fixed-1 is the worst strategy at every checkpoint, while full-pool population-based LLaMEA remains mid-table with an overall rank of $3.467$. The table also reports the original single-solution $(1+1)$-LLaMEA formulation as an additional reference to the original LLaMEA search setting; the main comparison instead uses the population-based extension described in~\Cref{app:frameworks}.

Across EoH and HiFo-Prompt frameworks, full-pool evaluation is consistently among the weakest strategies at intermediate checkpoints, whereas in LLaMEA it performs closer to the middle of the comparison. The preferred fixed evaluation depth also differs across frameworks: Fixed-25\% is the strongest fixed strategy overall under HiFo-Prompt, while Fixed-25\% and Fixed-50\% alternate as the strongest fixed strategies across the later EoH checkpoints. These differences reinforce that a single fixed evaluation depth does not transfer uniformly across search frameworks.

\begin{table}[t]
\centering
\caption{EoH anytime held-out performance, reported as mean rank (lower is better) at each search-budget checkpoint. Results are macro-averaged over the three CO benchmarks and two LLM backbones (Devstral and Qwen2.5-Coder). Bold and underline denote the best and runner-up strategies, respectively, in each column; higher values are better for the Best~\% and Top-2~\% columns.}
\label{tab:checkpoint_eoh_devstral_qwen}
\vspace{0.5em}
\begin{tabular}{lrrrrrrrr}
\toprule
 & \multicolumn{5}{c}{\textbf{Search Budget Consumed}} & \multicolumn{3}{c}{\textbf{Overall}} \\
\cmidrule(lr){2-6}\cmidrule(lr){7-9}
\textbf{Strategy} & 20\% & 40\% & 60\% & 80\% & 100\% & Rank $\downarrow$ & Best \% $\uparrow$ & Top-2 \% $\uparrow$ \\
\midrule
Fixed-1 & 3.528 & 3.389 & 3.792 & 4.000 & 4.208 & 3.783 & \underline{15.0} & 25.0 \\
Fixed-25\% & \textbf{2.583} & \underline{3.306} & 3.556 & 3.264 & 3.167 & \underline{3.175} & 11.7 & \underline{35.0} \\
Fixed-50\% & 3.611 & 3.458 & \underline{2.958} & \textbf{2.972} & \underline{2.944} & 3.189 & 10.6 & 27.2 \\
Full & 4.639 & 4.931 & 4.944 & 4.778 & 4.861 & 4.831 & 3.3 & 8.3 \\
AdaEva-S & 3.944 & 3.431 & 3.097 & \underline{3.014} & \textbf{2.625} & 3.222 & 8.9 & 27.8 \\
AdaEva-R & \underline{2.694} & \textbf{2.486} & \textbf{2.653} & \textbf{2.972} & 3.194 & \textbf{2.800} & \textbf{27.2} & \textbf{47.8} \\
\bottomrule
\end{tabular}
\end{table}

\begin{table}[t]
\centering
\caption{HiFo-Prompt anytime held-out performance, reported as mean rank (lower is better) at each search-budget checkpoint. Results are macro-averaged over the three CO benchmarks and two LLM backbones (Devstral and Qwen2.5-Coder). Bold and underline denote the best and runner-up strategies, respectively, in each column; higher values are better for the Best~\% and Top-2~\% columns.}
\label{tab:checkpoint_hifo_devstral_qwen}
\vspace{0.5em}
\begin{tabular}{lrrrrrrrr}
\toprule
 & \multicolumn{5}{c}{\textbf{Search Budget Consumed}} & \multicolumn{3}{c}{\textbf{Overall}} \\
\cmidrule(lr){2-6}\cmidrule(lr){7-9}
\textbf{Strategy} & 20\% & 40\% & 60\% & 80\% & 100\% & Rank $\downarrow$ & Best \% $\uparrow$ & Top-2 \% $\uparrow$ \\
\midrule
Fixed-1 & 4.472 & 4.639 & 4.556 & 4.639 & 4.681 & 4.597 & 6.7 & 12.8 \\
Fixed-25\% & 2.889 & \underline{2.708} & 2.806 & 3.014 & 2.986 & 2.881 & 16.7 & 32.8 \\
Fixed-50\% & 3.569 & 3.167 & 3.403 & 3.333 & 3.569 & 3.408 & 6.7 & 18.9 \\
Full & 4.889 & 5.097 & 5.111 & 5.000 & 4.569 & 4.933 & 1.7 & 5.0 \\
AdaEva-S & \textbf{2.472} & \textbf{2.625} & \underline{2.625} & \textbf{2.389} & \underline{2.681} & \textbf{2.558} & \underline{19.4} & \textbf{47.8} \\
AdaEva-R & \underline{2.708} & 2.764 & \textbf{2.500} & \underline{2.625} & \textbf{2.514} & \underline{2.622} & \textbf{20.6} & \underline{45.0} \\
\bottomrule
\end{tabular}
\end{table}

\begin{table}[t]
\centering
\caption{LLaMEA anytime held-out performance, reported as mean rank (lower is better) at each search-budget checkpoint. Results are macro-averaged over the 24 BBOB functions and two LLM backbones (Devstral and Qwen2.5-Coder), with the original single-solution $(1+1)$-LLaMEA configuration included as an additional reference. Bold and underline denote the best and runner-up strategies, respectively, in each column; higher values are better for the Best~\% and Top-2~\% columns.}
\label{tab:llamea_bbob_main}
\vspace{0.5em}
\begin{tabular}{lrrrrrrrr}
\toprule
 & \multicolumn{5}{c}{\textbf{Search Budget Consumed}} & \multicolumn{3}{c}{\textbf{Overall}} \\
\cmidrule(lr){2-6}\cmidrule(lr){7-9}
\textbf{Strategy} & 20\% & 40\% & 60\% & 80\% & 100\% & Rank $\downarrow$ & Best \% $\uparrow$ & Top-2 \% $\uparrow$ \\
\midrule
(1+1)-LLaMEA & 3.917 & 5.000 & 4.083 & 4.750 & 4.833 & 4.517 & 0.0 & 3.3 \\
(10+10)-LLaMEA (Full) & 3.917 & \underline{3.333} & 3.333 & 3.250 & 3.500 & 3.467 & 0.0 & 26.7 \\
Fixed-1 & 7.000 & 7.000 & 7.000 & 6.833 & 6.833 & 6.933 & 0.0 & 0.0 \\
Fixed-25\% & 4.500 & 3.667 & 4.500 & 4.583 & 4.750 & 4.400 & 0.0 & 13.3 \\
Fixed-50\% & 5.083 & 5.000 & 5.083 & 4.333 & 4.583 & 4.817 & 3.3 & 10.0 \\
AdaEva-S & \textbf{1.667} & \textbf{2.000} & \underline{2.167} & \underline{2.250} & \underline{2.000} & \underline{2.017} & \underline{46.7} & \underline{70.0} \\
AdaEva-R & \underline{1.917} & \textbf{2.000} & \textbf{1.833} & \textbf{2.000} & \textbf{1.500} & \textbf{1.850} & \textbf{50.0} & \textbf{76.7} \\
\bottomrule
\end{tabular}
\end{table}


\subsubsection{Objective Values per Instance Group}

\Cref{tab:eoh_obp_raw_results_devstral_p1,tab:eoh_obp_raw_results_devstral_p2} through~\Cref{tab:llamea_bbob_raw_results_devstral_p1,tab:llamea_bbob_raw_results_devstral_p2} report the underlying objective values for the Devstral backbone, broken down by held-out instance group, while~\Cref{tab:eoh_tsp_gemini} reports the corresponding EoH/TSP results with the Gemini-3.1-Flash-Lite backbone. These values provide the magnitude behind the rank comparisons. In particular, the tables show that the gaps between strategies can be small in FSSP even when their ranks are consistently separated, while larger differences emerge for some shifted OBP, higher-dimensional BBOB, and larger TSP instance groups.

For OBP, the capacity shift from $C=100$ to $C=500$ produces substantially larger gaps in the held-out objective than the in-distribution setting, particularly for smaller item pools. The relative ordering also becomes more variable under the shift: at the final checkpoint, AdaEva-S achieves the lowest objective across all three $C=500$ size groups, while its advantage is less uniform on the in-distribution groups. This illustrates that the adaptive strategies do not improve uniformly across every held-out condition, but can become particularly useful when the evaluation distribution differs from that used during search.

\begin{table}[t]
\centering
\small
\setlength{\tabcolsep}{5pt}
\caption{Anytime held-out performance of EoH on Online Bin Packing using the Devstral backbone. Cells report the incumbent's objective value at five search-budget checkpoints ($20\%$, $40\%$, $60\%$, $80\%$, and $100\%$), for in-distribution capacity $C{=}100$ and capacity-shifted $C{=}500$ across three instance sizes (1k, 5k, and 10k). Main entries are medians across five search seeds, with interquartile ranges shown in brackets. Lower values are better. (Part 1)}
\label{tab:eoh_obp_raw_results_devstral_p1}
\vspace{0.5em}
\begin{tabular}{llcccccc}
\toprule
& & \multicolumn{3}{c}{$C{=}100$ (in-distribution)} & \multicolumn{3}{c}{$C{=}500$ (capacity-shifted)} \\
\cmidrule(lr){3-5}\cmidrule(lr){6-8}
Budget & Method & 1k & 5k & 10k & 1k & 5k & 10k \\
\midrule
\multirow{12}{*}{20\%}
 & Fixed-1 & \textbf{3.59} & \textbf{1.09} & \textbf{0.56} & 6.31 & \textbf{1.83} & \textbf{1.23} \\
 & & \scriptsize[2.59, 4.33] & \scriptsize[0.69, 1.39] & \scriptsize[0.39, 1.01] & \scriptsize[4.82, 7.05] & \scriptsize[1.58, 2.43] & \scriptsize[0.93, 1.46] \\
 & Fixed-25\% & 3.84 & 2.23 & 1.83 & \textbf{4.57} & 3.13 & 2.60 \\
 & & \scriptsize[2.34, 4.58] & \scriptsize[1.78, 2.63] & \scriptsize[1.56, 2.05] & \scriptsize[3.08, 5.56] & \scriptsize[2.68, 3.28] & \scriptsize[2.30, 2.93] \\
 & Fixed-50\% & 4.83 & 2.18 & 1.36 & 12.02 & 5.61 & 4.19 \\
 & & \scriptsize[4.33, 5.57] & \scriptsize[1.69, 2.83] & \scriptsize[1.11, 1.71] & \scriptsize[11.03, 13.26] & \scriptsize[5.27, 6.16] & \scriptsize[3.67, 4.92] \\
 & Full & 4.83 & 3.87 & 3.32 & 6.06 & 4.67 & 3.50 \\
 & & \scriptsize[3.84, 5.82] & \scriptsize[3.48, 4.22] & \scriptsize[3.10, 3.77] & \scriptsize[4.82, 6.81] & \scriptsize[4.17, 4.92] & \scriptsize[3.35, 3.80] \\
 & AdaEva-S & 4.58 & 2.63 & 2.18 & 4.82 & 4.32 & 3.30 \\
 & & \scriptsize[3.34, 5.08] & \scriptsize[2.13, 3.03] & \scriptsize[2.00, 2.58] & \scriptsize[3.58, 6.06] & \scriptsize[4.02, 4.67] & \scriptsize[3.03, 3.50] \\
 & AdaEva-R & 4.83 & 2.13 & 1.28 & 12.52 & 4.47 & 4.19 \\
 & & \scriptsize[3.59, 5.57] & \scriptsize[1.78, 3.23] & \scriptsize[0.88, 1.61] & \scriptsize[11.77, 13.51] & \scriptsize[4.22, 5.22] & \scriptsize[3.97, 4.44] \\
\cmidrule(lr){1-8}
\multirow{12}{*}{40\%}
 & Fixed-1 & 4.08 & \textbf{1.09} & \textbf{0.51} & 12.77 & 4.07 & 2.33 \\
 & & \scriptsize[3.59, 4.58] & \scriptsize[0.69, 1.54] & \scriptsize[0.39, 0.78] & \scriptsize[11.77, 14.01] & \scriptsize[3.67, 4.82] & \scriptsize[2.15, 2.60] \\
 & Fixed-25\% & \textbf{3.09} & 1.29 & 0.83 & 8.54 & \textbf{2.98} & \textbf{1.66} \\
 & & \scriptsize[2.10, 4.33] & \scriptsize[0.79, 1.49] & \scriptsize[0.64, 1.01] & \scriptsize[7.80, 10.28] & \scriptsize[2.43, 3.33] & \scriptsize[1.46, 1.93] \\
 & Fixed-50\% & 3.84 & 1.39 & 0.86 & 11.03 & 4.42 & 2.43 \\
 & & \scriptsize[3.09, 4.83] & \scriptsize[0.84, 1.69] & \scriptsize[0.68, 1.06] & \scriptsize[9.79, 12.02] & \scriptsize[3.97, 5.02] & \scriptsize[2.23, 2.88] \\
 & Full & 6.07 & 3.33 & 2.82 & 14.51 & 8.10 & 5.59 \\
 & & \scriptsize[4.83, 6.32] & \scriptsize[2.93, 3.87] & \scriptsize[2.60, 3.17] & \scriptsize[13.51, 15.75] & \scriptsize[7.65, 8.65] & \scriptsize[5.07, 5.99] \\
 & AdaEva-S & 4.58 & 2.13 & 1.56 & \textbf{5.07} & 4.22 & 3.30 \\
 & & \scriptsize[3.34, 5.08] & \scriptsize[1.78, 2.83] & \scriptsize[1.46, 1.80] & \scriptsize[3.58, 5.81] & \scriptsize[3.97, 4.67] & \scriptsize[2.85, 3.50] \\
 & AdaEva-R & 4.33 & 1.59 & 0.78 & 13.02 & 4.87 & 2.78 \\
 & & \scriptsize[3.59, 5.08] & \scriptsize[1.14, 1.88] & \scriptsize[0.61, 1.03] & \scriptsize[12.27, 14.75] & \scriptsize[4.47, 5.22] & \scriptsize[2.45, 3.20] \\
\bottomrule
\end{tabular}
\end{table}

For TSP, performance degrades systematically as the held-out instance size increases from the in-distribution group ($\leq 100$ cities) to $101$--$150$ and then $>150$ cities. These extrapolation groups therefore provide a more demanding test for all strategies. AdaEva-R is particularly competitive on the larger groups at intermediate checkpoints, while the differences among strategies are smaller on the in-distribution group at later checkpoints.

For FSSP, the objective differences are comparatively small on the $m=5$ group, where most methods achieve nearly identical performance. Differences become more visible as the number of machines increases, especially for $m=20$, although the absolute gaps remain modest. AdaEva-R is consistently competitive on the larger machine groups, while AdaEva-S also becomes increasingly competitive as the search progresses.

For BBOB, the higher-dimensional held-out settings are more difficult for all methods, with AOCC generally decreasing as the dimension increases from $5$ to $20$. The adaptive methods maintain higher AOCC than the fixed-evaluation strategies across dimensions, with AdaEva-S particularly strong at dimensions $10$ and $20$ and AdaEva-R strongest overall at several checkpoints. Thus, the advantage observed in the rank analysis is also visible in the underlying objective values rather than arising solely from rank aggregation.

\begin{table}[t]
\centering
\small
\setlength{\tabcolsep}{5pt}
\caption{Anytime held-out performance of EoH on Online Bin Packing using the Devstral backbone. Cells report the incumbent's objective value at five search-budget checkpoints ($20\%$, $40\%$, $60\%$, $80\%$, and $100\%$), for in-distribution capacity $C{=}100$ and capacity-shifted $C{=}500$ across three instance sizes (1k, 5k, and 10k). Main entries are medians across five search seeds, with interquartile ranges shown in brackets. Lower values are better. (Part 2)}
\label{tab:eoh_obp_raw_results_devstral_p2}
\vspace{0.5em}
\begin{tabular}{llcccccc}
\toprule
& & \multicolumn{3}{c}{$C{=}100$ (in-distribution)} & \multicolumn{3}{c}{$C{=}500$ (capacity-shifted)} \\
\cmidrule(lr){3-5}\cmidrule(lr){6-8}
Budget & Method & 1k & 5k & 10k & 1k & 5k & 10k \\
\midrule
\multirow{12}{*}{60\%}
 & Fixed-1 & 4.08 & 1.19 & \textbf{0.54} & 12.77 & 3.97 & 2.33 \\
 & & \scriptsize[3.09, 4.83] & \scriptsize[0.94, 1.69] & \scriptsize[0.46, 1.06] & \scriptsize[11.77, 13.51] & \scriptsize[3.67, 4.67] & \scriptsize[2.15, 2.60] \\
 & Fixed-25\% & \textbf{3.34} & \textbf{1.09} & 0.59 & 10.53 & \textbf{3.42} & \textbf{2.10} \\
 & & \scriptsize[2.10, 3.84] & \scriptsize[0.64, 1.49] & \scriptsize[0.41, 0.88] & \scriptsize[9.04, 11.53] & \scriptsize[2.98, 3.87] & \scriptsize[1.78, 2.43] \\
 & Fixed-50\% & 4.08 & 1.19 & 0.86 & \textbf{9.54} & 3.87 & 2.18 \\
 & & \scriptsize[2.84, 4.83] & \scriptsize[0.94, 1.59] & \scriptsize[0.73, 1.16] & \scriptsize[8.79, 11.53] & \scriptsize[3.42, 4.67] & \scriptsize[1.93, 2.35] \\
 & Full & 4.33 & 2.78 & 1.95 & 10.53 & 4.07 & 3.45 \\
 & & \scriptsize[3.84, 5.08] & \scriptsize[2.33, 3.08] & \scriptsize[1.66, 2.20] & \scriptsize[9.79, 11.77] & \scriptsize[3.67, 4.52] & \scriptsize[3.22, 3.67] \\
 & AdaEva-S & 4.08 & 1.54 & 0.96 & 12.52 & 4.47 & 3.10 \\
 & & \scriptsize[3.59, 5.08] & \scriptsize[0.99, 1.93] & \scriptsize[0.68, 1.11] & \scriptsize[11.53, 13.76] & \scriptsize[3.97, 5.22] & \scriptsize[2.68, 3.22] \\
 & AdaEva-R & 4.58 & 1.49 & 0.78 & 13.26 & 4.42 & 2.53 \\
 & & \scriptsize[3.59, 5.08] & \scriptsize[0.89, 1.78] & \scriptsize[0.59, 1.11] & \scriptsize[12.27, 14.51] & \scriptsize[4.07, 5.22] & \scriptsize[2.30, 2.93] \\
\cmidrule(lr){1-8}
\multirow{12}{*}{80\%}
 & Fixed-1 & 3.84 & 1.19 & 0.61 & 12.52 & 3.87 & 2.13 \\
 & & \scriptsize[3.09, 4.58] & \scriptsize[0.94, 1.59] & \scriptsize[0.49, 1.06] & \scriptsize[11.53, 13.51] & \scriptsize[3.47, 4.32] & \scriptsize[1.88, 2.48] \\
 & Fixed-25\% & \textbf{3.09} & \textbf{1.09} & \textbf{0.59} & 11.03 & 3.57 & \textbf{1.93} \\
 & & \scriptsize[1.85, 3.84] & \scriptsize[0.64, 1.34] & \scriptsize[0.41, 0.88] & \scriptsize[9.79, 11.77] & \scriptsize[2.98, 3.97] & \scriptsize[1.66, 2.35] \\
 & Fixed-50\% & 3.59 & \textbf{1.09} & 0.64 & 12.77 & 3.67 & 2.13 \\
 & & \scriptsize[2.34, 4.33] & \scriptsize[0.69, 1.39] & \scriptsize[0.46, 0.86] & \scriptsize[11.77, 14.01] & \scriptsize[3.42, 4.47] & \scriptsize[1.86, 2.30] \\
 & Full & 4.08 & 2.68 & 1.88 & \textbf{6.56} & \textbf{3.28} & 3.07 \\
 & & \scriptsize[3.34, 5.08] & \scriptsize[2.33, 3.03] & \scriptsize[1.56, 2.03] & \scriptsize[4.82, 7.55] & \scriptsize[2.98, 3.57] & \scriptsize[2.88, 3.25] \\
 & AdaEva-S & 3.84 & 1.14 & 0.68 & 13.76 & 3.97 & 2.15 \\
 & & \scriptsize[2.84, 4.33] & \scriptsize[0.79, 1.59] & \scriptsize[0.41, 0.93] & \scriptsize[12.27, 14.75] & \scriptsize[3.67, 4.77] & \scriptsize[2.03, 2.45] \\
 & AdaEva-R & 4.33 & 1.34 & 0.76 & 13.51 & 4.87 & 2.68 \\
 & & \scriptsize[3.59, 5.33] & \scriptsize[0.94, 1.69] & \scriptsize[0.51, 1.01] & \scriptsize[12.52, 15.00] & \scriptsize[4.32, 5.41] & \scriptsize[2.40, 3.07] \\
\cmidrule(lr){1-8}
\multirow{12}{*}{100\%}
 & Fixed-1 & 3.84 & 1.29 & 0.61 & 12.52 & 3.87 & 2.13 \\
 & & \scriptsize[3.09, 4.58] & \scriptsize[0.94, 1.59] & \scriptsize[0.49, 1.06] & \scriptsize[11.53, 13.51] & \scriptsize[3.47, 4.32] & \scriptsize[1.88, 2.48] \\
 & Fixed-25\% & \textbf{2.84} & 1.04 & 0.61 & 10.28 & 3.42 & 1.91 \\
 & & \scriptsize[1.85, 3.84] & \scriptsize[0.64, 1.29] & \scriptsize[0.41, 0.88] & \scriptsize[9.29, 11.77] & \scriptsize[2.98, 3.77] & \scriptsize[1.63, 2.30] \\
 & Fixed-50\% & 3.34 & \textbf{0.89} & \textbf{0.56} & 12.52 & 3.67 & 1.91 \\
 & & \scriptsize[2.34, 4.33] & \scriptsize[0.49, 1.24] & \scriptsize[0.39, 0.81] & \scriptsize[12.02, 14.01] & \scriptsize[3.33, 4.27] & \scriptsize[1.68, 2.18] \\
 & Full & 4.33 & 1.74 & 1.18 & 12.02 & 4.37 & 2.85 \\
 & & \scriptsize[3.34, 5.08] & \scriptsize[1.49, 2.23] & \scriptsize[1.06, 1.51] & \scriptsize[11.28, 12.77] & \scriptsize[3.92, 4.92] & \scriptsize[2.65, 3.10] \\
 & AdaEva-S & \textbf{2.84} & 1.14 & 0.66 & \textbf{8.54} & \textbf{2.58} & \textbf{1.48} \\
 & & \scriptsize[1.85, 3.84] & \scriptsize[0.74, 1.49] & \scriptsize[0.46, 0.88] & \scriptsize[7.55, 9.79] & \scriptsize[2.18, 3.23] & \scriptsize[1.18, 1.68] \\
 & AdaEva-R & 4.08 & 1.34 & 0.68 & 13.51 & 4.32 & 2.60 \\
 & & \scriptsize[3.34, 4.83] & \scriptsize[0.79, 1.59] & \scriptsize[0.49, 1.01] & \scriptsize[12.27, 14.75] & \scriptsize[4.07, 5.02] & \scriptsize[2.30, 2.78] \\
\bottomrule
\end{tabular}
\end{table}

\begin{table}[t]
\centering
\small
\caption{Anytime held-out performance of EoH on TSPLIB using the Devstral backbone. Cells report the incumbent's objective value at five search-budget checkpoints ($20\%$, $40\%$, $60\%$, $80\%$, and $100\%$), for three instance-size groups: $\leq 100$ cities (12 in-distribution instances), $101$--$150$ cities (11 instances), and $>150$ cities (6 extrapolation instances). Main entries are medians across five search seeds, with interquartile ranges shown in brackets. Lower values are better. (Part 1)}
\vspace{0.5em}
\label{tab:eoh_tsp_raw_results_devstral_p1}
\begin{tabular}{llccc}
\toprule
& & $\leq 100$ (in-distribution) & $101$--$150$ & $>150$ (extrapolation) \\
\cmidrule(lr){3-3}\cmidrule(lr){4-4}\cmidrule(lr){5-5}
Budget & Method & 12 instances & 11 instances & 6 instances \\
\midrule
\multirow{12}{*}{20\%}
 & Fixed-1 & 0.034 & 0.168 & 0.694 \\
 & & \scriptsize[0.008, 0.406] & \scriptsize[0.015, 0.370] & \scriptsize[0.159, 0.906] \\
 & Fixed-25\% & 0.028 & 0.318 & \textbf{0.521} \\
 & & \scriptsize[0.008, 0.403] & \scriptsize[0.015, 0.499] & \scriptsize[0.124, 0.988] \\
 & Fixed-50\% & 0.244 & 0.744 & 0.873 \\
 & & \scriptsize[0.026, 0.554] & \scriptsize[0.258, 1.359] & \scriptsize[0.827, 1.393] \\
 & Full & 4.020 & 4.194 & 2.083 \\
 & & \scriptsize[1.836, 5.801] & \scriptsize[2.115, 6.330] & \scriptsize[1.645, 4.086] \\
 & AdaEva-S & 1.898 & 1.789 & 1.449 \\
 & & \scriptsize[0.929, 2.384] & \scriptsize[0.752, 3.003] & \scriptsize[1.002, 3.244] \\
 & AdaEva-R & \textbf{0.020} & \textbf{0.110} & 0.545 \\
 & & \scriptsize[0.004, 0.403] & \scriptsize[0.002, 0.270] & \scriptsize[0.051, 1.125] \\
\cmidrule(lr){1-5}
\multirow{12}{*}{40\%}
 & Fixed-1 & 0.059 & 0.203 & 0.533 \\
 & & \scriptsize[0.008, 0.403] & \scriptsize[0.015, 0.381] & \scriptsize[0.132, 0.685] \\
 & Fixed-25\% & \textbf{0.024} & 0.148 & 0.427 \\
 & & \scriptsize[0.004, 0.403] & \scriptsize[0.015, 0.380] & \scriptsize[0.108, 0.500] \\
 & Fixed-50\% & 0.071 & \textbf{0.093} & 0.471 \\
 & & \scriptsize[0.007, 0.403] & \scriptsize[0.007, 0.360] & \scriptsize[0.102, 1.170] \\
 & Full & 1.717 & 1.269 & 1.237 \\
 & & \scriptsize[0.924, 2.767] & \scriptsize[0.670, 4.543] & \scriptsize[0.986, 3.476] \\
 & AdaEva-S & \textbf{0.024} & 0.239 & 0.492 \\
 & & \scriptsize[0.005, 0.403] & \scriptsize[0.015, 0.504] & \scriptsize[0.099, 0.861] \\
 & AdaEva-R & \textbf{0.024} & 0.102 & \textbf{0.309} \\
 & & \scriptsize[0.004, 0.403] & \scriptsize[0.015, 0.288] & \scriptsize[0.070, 0.398] \\
\bottomrule
\end{tabular}
\end{table}

\begin{table}[t]
\centering
\small
\caption{Anytime held-out performance of EoH on TSPLIB using the Devstral backbone. Cells report the incumbent's objective value at five search-budget checkpoints ($20\%$, $40\%$, $60\%$, $80\%$, and $100\%$), for three instance-size groups: $\leq 100$ cities (12 in-distribution instances), $101$--$150$ cities (11 instances), and $>150$ cities (6 extrapolation instances). Main entries are medians across five search seeds, with interquartile ranges shown in brackets. Lower values are better. (Part 2)}
\vspace{0.5em}
\label{tab:eoh_tsp_raw_results_devstral_p2}
\begin{tabular}{llccc}
\toprule
& & $\leq 100$ (in-distribution) & $101$--$150$ & $>150$ (extrapolation) \\
\cmidrule(lr){3-3}\cmidrule(lr){4-4}\cmidrule(lr){5-5}
Budget & Method & 12 instances & 11 instances & 6 instances \\
\midrule
\multirow{12}{*}{60\%}
 & Fixed-1 & 0.059 & 0.203 & 0.533 \\
 & & \scriptsize[0.008, 0.403] & \scriptsize[0.015, 0.381] & \scriptsize[0.132, 0.685] \\
 & Fixed-25\% & 0.024 & 0.244 & 0.486 \\
 & & \scriptsize[0.004, 0.403] & \scriptsize[0.077, 0.526] & \scriptsize[0.102, 0.705] \\
 & Fixed-50\% & \textbf{0.012} & 0.085 & 0.399 \\
 & & \scriptsize[0.003, 0.403] & \scriptsize[0.015, 0.367] & \scriptsize[0.097, 0.579] \\
 & Full & 1.070 & 1.340 & 1.642 \\
 & & \scriptsize[0.681, 2.158] & \scriptsize[1.130, 3.011] & \scriptsize[1.277, 3.203] \\
 & AdaEva-S & 0.020 & 0.147 & 0.339 \\
 & & \scriptsize[0.004, 0.403] & \scriptsize[0.015, 0.215] & \scriptsize[0.074, 0.764] \\
 & AdaEva-R & 0.020 & \textbf{0.048} & \textbf{0.197} \\
 & & \scriptsize[0.004, 0.403] & \scriptsize[0.015, 0.286] & \scriptsize[0.028, 0.489] \\
\cmidrule(lr){1-5}
\multirow{12}{*}{80\%}
 & Fixed-1 & 0.059 & 0.203 & 0.533 \\
 & & \scriptsize[0.008, 0.403] & \scriptsize[0.015, 0.381] & \scriptsize[0.132, 0.685] \\
 & Fixed-25\% & 0.020 & 0.045 & 0.421 \\
 & & \scriptsize[0.004, 0.403] & \scriptsize[0.003, 0.091] & \scriptsize[0.105, 0.435] \\
 & Fixed-50\% & 0.021 & 0.100 & 0.365 \\
 & & \scriptsize[0.004, 0.403] & \scriptsize[0.002, 0.345] & \scriptsize[0.053, 0.528] \\
 & Full & 1.144 & 0.971 & 1.262 \\
 & & \scriptsize[0.978, 1.391] & \scriptsize[0.621, 1.949] & \scriptsize[0.921, 1.609] \\
 & AdaEva-S & \textbf{0.012} & \textbf{0.033} & 0.305 \\
 & & \scriptsize[0.003, 0.403] & \scriptsize[0.008, 0.170] & \scriptsize[0.064, 0.428] \\
 & AdaEva-R & \textbf{0.012} & 0.048 & \textbf{0.242} \\
 & & \scriptsize[0.003, 0.403] & \scriptsize[0.015, 0.267] & \scriptsize[0.056, 0.322] \\
\cmidrule(lr){1-5}
\multirow{12}{*}{100\%}
 & Fixed-1 & 0.059 & 0.203 & 0.533 \\
 & & \scriptsize[0.008, 0.403] & \scriptsize[0.015, 0.381] & \scriptsize[0.132, 0.685] \\
 & Fixed-25\% & 0.024 & 0.045 & 0.421 \\
 & & \scriptsize[0.004, 0.403] & \scriptsize[0.003, 0.091] & \scriptsize[0.105, 0.435] \\
 & Fixed-50\% & 0.021 & 0.050 & 0.365 \\
 & & \scriptsize[0.004, 0.403] & \scriptsize[0.002, 0.345] & \scriptsize[0.053, 0.528] \\
 & Full & 0.159 & 0.550 & 0.864 \\
 & & \scriptsize[0.016, 0.406] & \scriptsize[0.025, 0.698] & \scriptsize[0.382, 1.236] \\
 & AdaEva-S & 0.024 & \textbf{0.033} & \textbf{0.305} \\
 & & \scriptsize[0.004, 0.403] & \scriptsize[0.008, 0.170] & \scriptsize[0.064, 0.428] \\
 & AdaEva-R & \textbf{0.012} & 0.048 & 0.306 \\
 & & \scriptsize[0.003, 0.403] & \scriptsize[0.015, 0.267] & \scriptsize[0.070, 0.495] \\
\bottomrule
\end{tabular}
\end{table}

\begin{table}[t]
\centering
\small
\caption{Anytime held-out performance of EoH on the Taillard flow-shop scheduling benchmark using the Devstral backbone. Cells report the incumbent's objective value at five search-budget checkpoints ($20\%$, $40\%$, $60\%$, $80\%$, and $100\%$), for three problem-size groups: $m{=}5$ (30 instances), $m{=}10$ (40 instances), and $m{=}20$ (40 instances). Main entries are medians across five search seeds, with interquartile ranges shown in brackets. Lower values are better. (Part 1)}
\vspace{0.5em}
\label{tab:eoh_fssp_raw_results_devstral_p1}
\begin{tabular}{llccc}
\toprule
Budget & Method & $m{=}5$ (30) & $m{=}10$ (40) & $m{=}20$ (40) \\
\midrule
\multirow{12}{*}{20\%}
 & Fixed-1 & \textbf{0.00} & 0.51 & 1.49 \\
 & & \scriptsize[0.00, 0.07] & \scriptsize[0.29, 0.81] & \scriptsize[1.01, 1.86] \\
 & Fixed-25\% & \textbf{0.00} & \textbf{0.32} & 1.34 \\
 & & \scriptsize[0.00, 0.00] & \scriptsize[0.22, 0.57] & \scriptsize[0.73, 1.57] \\
 & Fixed-50\% & \textbf{0.00} & 0.43 & 1.39 \\
 & & \scriptsize[0.00, 0.04] & \scriptsize[0.24, 0.66] & \scriptsize[0.80, 1.62] \\
 & Full & \textbf{0.00} & 0.49 & 1.26 \\
 & & \scriptsize[0.00, 0.10] & \scriptsize[0.22, 0.76] & \scriptsize[0.74, 1.61] \\
 & AdaEva-S & \textbf{0.00} & 0.49 & 1.29 \\
 & & \scriptsize[0.00, 0.10] & \scriptsize[0.22, 0.76] & \scriptsize[0.69, 1.61] \\
 & AdaEva-R & \textbf{0.00} & 0.39 & \textbf{1.22} \\
 & & \scriptsize[0.00, 0.00] & \scriptsize[0.23, 0.58] & \scriptsize[0.53, 1.46] \\
\cmidrule(lr){1-5}
\multirow{12}{*}{40\%}
 & Fixed-1 & \textbf{0.00} & 0.43 & 1.43 \\
 & & \scriptsize[0.00, 0.07] & \scriptsize[0.25, 0.66] & \scriptsize[0.84, 1.86] \\
 & Fixed-25\% & \textbf{0.00} & 0.35 & 1.28 \\
 & & \scriptsize[0.00, 0.00] & \scriptsize[0.19, 0.56] & \scriptsize[0.66, 1.61] \\
 & Fixed-50\% & \textbf{0.00} & 0.37 & 1.44 \\
 & & \scriptsize[0.00, 0.05] & \scriptsize[0.24, 0.57] & \scriptsize[0.78, 1.62] \\
 & Full & \textbf{0.00} & 0.49 & 1.26 \\
 & & \scriptsize[0.00, 0.10] & \scriptsize[0.22, 0.76] & \scriptsize[0.69, 1.61] \\
 & AdaEva-S & \textbf{0.00} & 0.41 & 1.27 \\
 & & \scriptsize[0.00, 0.00] & \scriptsize[0.14, 0.49] & \scriptsize[0.65, 1.59] \\
 & AdaEva-R & \textbf{0.00} & \textbf{0.33} & \textbf{1.22} \\
 & & \scriptsize[0.00, 0.00] & \scriptsize[0.12, 0.55] & \scriptsize[0.76, 1.46] \\
\cmidrule(lr){1-5}
\multirow{12}{*}{60\%}
 & Fixed-1 & \textbf{0.00} & 0.43 & 1.43 \\
 & & \scriptsize[0.00, 0.07] & \scriptsize[0.25, 0.66] & \scriptsize[0.84, 1.86] \\
 & Fixed-25\% & \textbf{0.00} & 0.39 & 1.22 \\
 & & \scriptsize[0.00, 0.00] & \scriptsize[0.12, 0.56] & \scriptsize[0.73, 1.62] \\
 & Fixed-50\% & \textbf{0.00} & 0.41 & 1.39 \\
 & & \scriptsize[0.00, 0.00] & \scriptsize[0.24, 0.56] & \scriptsize[0.68, 1.63] \\
 & Full & \textbf{0.00} & 0.49 & 1.26 \\
 & & \scriptsize[0.00, 0.10] & \scriptsize[0.22, 0.76] & \scriptsize[0.69, 1.61] \\
 & AdaEva-S & \textbf{0.00} & 0.35 & 1.35 \\
 & & \scriptsize[0.00, 0.00] & \scriptsize[0.12, 0.48] & \scriptsize[0.79, 1.55] \\
 & AdaEva-R & \textbf{0.00} & \textbf{0.32} & \textbf{1.19} \\
 & & \scriptsize[0.00, 0.00] & \scriptsize[0.12, 0.55] & \scriptsize[0.70, 1.48] \\
\cmidrule(lr){1-5}
\multirow{12}{*}{80\%}
 & Fixed-1 & \textbf{0.00} & 0.43 & 1.43 \\
 & & \scriptsize[0.00, 0.07] & \scriptsize[0.25, 0.66] & \scriptsize[0.84, 1.87] \\
 & Fixed-25\% & \textbf{0.00} & 0.39 & 1.23 \\
 & & \scriptsize[0.00, 0.00] & \scriptsize[0.15, 0.58] & \scriptsize[0.71, 1.62] \\
 & Fixed-50\% & \textbf{0.00} & 0.36 & 1.33 \\
 & & \scriptsize[0.00, 0.00] & \scriptsize[0.21, 0.60] & \scriptsize[0.72, 1.61] \\
 & Full & \textbf{0.00} & 0.42 & 1.27 \\
 & & \scriptsize[0.00, 0.04] & \scriptsize[0.21, 0.59] & \scriptsize[0.67, 1.61] \\
 & AdaEva-S & \textbf{0.00} & 0.34 & 1.22 \\
 & & \scriptsize[0.00, 0.00] & \scriptsize[0.10, 0.64] & \scriptsize[0.51, 1.43] \\
 & AdaEva-R & \textbf{0.00} & \textbf{0.33} & \textbf{1.14} \\
 & & \scriptsize[0.00, 0.00] & \scriptsize[0.19, 0.52] & \scriptsize[0.67, 1.47] \\
\bottomrule
\end{tabular}
\end{table}

\begin{table}[t]
\centering
\small
\caption{Anytime held-out performance of EoH on the Taillard flow-shop scheduling benchmark using the Devstral backbone. Cells report the incumbent's objective value at five search-budget checkpoints ($20\%$, $40\%$, $60\%$, $80\%$, and $100\%$), for three problem-size groups: $m{=}5$ (30 instances), $m{=}10$ (40 instances), and $m{=}20$ (40 instances). Main entries are medians across five search seeds, with interquartile ranges shown in brackets. Lower values are better. The lower block reports the percentage of instances whose normalized gap is at or below the upper bound (gap $\leq 0$). (Part 2)}
\vspace{0.5em}
\label{tab:eoh_fssp_raw_results_devstral_p2}
\begin{tabular}{llccc}
\toprule
Budget & Method & $m{=}5$ (30) & $m{=}10$ (40) & $m{=}20$ (40) \\
\midrule
\multirow{12}{*}{100\%}
 & Fixed-1 & \textbf{0.00} & 0.43 & 1.43 \\
 & & \scriptsize[0.00, 0.07] & \scriptsize[0.25, 0.66] & \scriptsize[0.84, 1.83] \\
 & Fixed-25\% & \textbf{0.00} & 0.41 & 1.22 \\
 & & \scriptsize[0.00, 0.00] & \scriptsize[0.18, 0.58] & \scriptsize[0.71, 1.56] \\
 & Fixed-50\% & \textbf{0.00} & 0.32 & 1.23 \\
 & & \scriptsize[0.00, 0.00] & \scriptsize[0.21, 0.61] & \scriptsize[0.57, 1.60] \\
 & Full & \textbf{0.00} & 0.35 & 1.29 \\
 & & \scriptsize[0.00, 0.05] & \scriptsize[0.24, 0.57] & \scriptsize[0.79, 1.59] \\
 & AdaEva-S & \textbf{0.00} & \textbf{0.30} & 1.24 \\
 & & \scriptsize[0.00, 0.00] & \scriptsize[0.09, 0.55] & \scriptsize[0.49, 1.47] \\
 & AdaEva-R & \textbf{0.00} & 0.33 & \textbf{1.13} \\
 & & \scriptsize[0.00, 0.00] & \scriptsize[0.21, 0.59] & \scriptsize[0.69, 1.44] \\
\midrule
\multicolumn{5}{l}{\emph{Instances at or below the upper bound (gap $\leq 0$)}} \\
\midrule
\multirow{6}{*}{20\%}
 & Fixed-1 & 67\% & 8\% & 2\% \\
 & Fixed-25\% & 77\% & \textbf{10\%} & 5\% \\
 & Fixed-50\% & 73\% & 5\% & 2\% \\
 & Full & 60\% & 8\% & 2\% \\
 & AdaEva-S & 60\% & 8\% & 2\% \\
 & AdaEva-R & \textbf{80\%} & 8\% & \textbf{10\%} \\
\cmidrule(lr){1-5}
\multirow{6}{*}{40\%}
 & Fixed-1 & 70\% & 10\% & 0\% \\
 & Fixed-25\% & 80\% & 10\% & 5\% \\
 & Fixed-50\% & 70\% & 5\% & 5\% \\
 & Full & 60\% & 8\% & 2\% \\
 & AdaEva-S & 83\% & 8\% & \textbf{8\%} \\
 & AdaEva-R & \textbf{87\%} & \textbf{15\%} & \textbf{8\%} \\
\cmidrule(lr){1-5}
\multirow{6}{*}{60\%}
 & Fixed-1 & 70\% & \textbf{10\%} & 0\% \\
 & Fixed-25\% & \textbf{83\%} & \textbf{10\%} & \textbf{12\%} \\
 & Fixed-50\% & 77\% & \textbf{10\%} & 2\% \\
 & Full & 60\% & 8\% & 2\% \\
 & AdaEva-S & \textbf{83\%} & \textbf{10\%} & 8\% \\
 & AdaEva-R & \textbf{83\%} & \textbf{10\%} & 8\% \\
\cmidrule(lr){1-5}
\multirow{6}{*}{80\%}
 & Fixed-1 & 70\% & 10\% & 0\% \\
 & Fixed-25\% & 87\% & \textbf{12\%} & \textbf{12\%} \\
 & Fixed-50\% & 87\% & 5\% & 10\% \\
 & Full & 73\% & 8\% & 0\% \\
 & AdaEva-S & 80\% & 10\% & 5\% \\
 & AdaEva-R & \textbf{90\%} & 5\% & 8\% \\
\cmidrule(lr){1-5}
\multirow{6}{*}{100\%}
 & Fixed-1 & 70\% & 10\% & 0\% \\
 & Fixed-25\% & 83\% & \textbf{12\%} & 5\% \\
 & Fixed-50\% & 87\% & 8\% & \textbf{8\%} \\
 & Full & 70\% & 8\% & 2\% \\
 & AdaEva-S & 87\% & \textbf{12\%} & 2\% \\
 & AdaEva-R & \textbf{90\%} & 8\% & \textbf{8\%} \\
\bottomrule
\end{tabular}
\end{table}

\begin{table}[t]
\centering
\small
\caption{Anytime held-out performance of LLaMEA on BBOB using the Devstral backbone. Cells report AOCC at five search-budget checkpoints ($20\%$, $40\%$, $60\%$, $80\%$, and $100\%$), evaluated on held-out instances across all 24 BBOB functions and dimensions $5$, $10$, and $20$. \emph{All} denotes the mean across the three dimensions. Main entries are medians across five search seeds, with interquartile ranges shown in brackets. Higher AOCC is better. (Part 1)}
\vspace{0.5em}
\label{tab:llamea_bbob_raw_results_devstral_p1}
\begin{tabular}{llcccc}
\toprule
Budget & Method & $d{=}5$ & $d{=}10$ & $d{=}20$ & All \\
\midrule
\multirow{14}{*}{20\%}
 & (1+1)-LLaMEA & 0.505 & 0.458 & 0.429 & 0.464 \\
 & & \scriptsize[0.477, 0.517] & \scriptsize[0.435, 0.463] & \scriptsize[0.401, 0.435] & \scriptsize[0.438, 0.472] \\
 & (10+10)-LLaMEA & 0.526 & 0.471 & 0.428 & 0.474 \\
 & & \scriptsize[0.517, 0.543] & \scriptsize[0.452, 0.474] & \scriptsize[0.389, 0.431] & \scriptsize[0.462, 0.475] \\
 & Fixed-1 & 0.486 & 0.433 & 0.402 & 0.439 \\
 & & \scriptsize[0.458, 0.521] & \scriptsize[0.424, 0.459] & \scriptsize[0.400, 0.418] & \scriptsize[0.428, 0.464] \\
 & Fixed-25\% & 0.535 & 0.460 & 0.427 & 0.464 \\
 & & \scriptsize[0.514, 0.539] & \scriptsize[0.457, 0.483] & \scriptsize[0.387, 0.458] & \scriptsize[0.459, 0.485] \\
 & Fixed-50\% & 0.548 & 0.460 & 0.417 & 0.469 \\
 & & \scriptsize[0.520, 0.556] & \scriptsize[0.456, 0.465] & \scriptsize[0.385, 0.421] & \scriptsize[0.462, 0.478] \\
 & AdaEva-S & 0.543 & 0.472 & \textbf{0.438} & 0.477 \\
 & & \scriptsize[0.531, 0.546] & \scriptsize[0.428, 0.505] & \scriptsize[0.363, 0.466] & \scriptsize[0.444, 0.510] \\
 & AdaEva-R & \textbf{0.584} & \textbf{0.487} & 0.427 & \textbf{0.499} \\
 & & \scriptsize[0.582, 0.586] & \scriptsize[0.465, 0.495] & \scriptsize[0.398, 0.448] & \scriptsize[0.483, 0.510] \\
\cmidrule(lr){1-6}
\multirow{14}{*}{40\%}
 & (1+1)-LLaMEA & 0.513 & 0.460 & 0.428 & 0.467 \\
 & & \scriptsize[0.476, 0.530] & \scriptsize[0.431, 0.470] & \scriptsize[0.397, 0.438] & \scriptsize[0.435, 0.480] \\
 & (10+10)-LLaMEA & 0.540 & 0.489 & 0.429 & 0.488 \\
 & & \scriptsize[0.532, 0.543] & \scriptsize[0.483, 0.494] & \scriptsize[0.405, 0.437] & \scriptsize[0.480, 0.496] \\
 & Fixed-1 & 0.480 & 0.434 & 0.403 & 0.438 \\
 & & \scriptsize[0.459, 0.513] & \scriptsize[0.425, 0.460] & \scriptsize[0.401, 0.420] & \scriptsize[0.429, 0.463] \\
 & Fixed-25\% & 0.561 & 0.483 & 0.411 & 0.485 \\
 & & \scriptsize[0.549, 0.565] & \scriptsize[0.465, 0.487] & \scriptsize[0.390, 0.458] & \scriptsize[0.474, 0.486] \\
 & Fixed-50\% & 0.548 & 0.463 & 0.405 & 0.467 \\
 & & \scriptsize[0.520, 0.557] & \scriptsize[0.430, 0.465] & \scriptsize[0.366, 0.426] & \scriptsize[0.451, 0.484] \\
 & AdaEva-S & 0.546 & \textbf{0.512} & \textbf{0.478} & \textbf{0.515} \\
 & & \scriptsize[0.546, 0.553] & \scriptsize[0.499, 0.517] & \scriptsize[0.444, 0.497] & \scriptsize[0.496, 0.522] \\
 & AdaEva-R & \textbf{0.590} & 0.488 & 0.427 & 0.508 \\
 & & \scriptsize[0.586, 0.592] & \scriptsize[0.487, 0.497] & \scriptsize[0.420, 0.445] & \scriptsize[0.500, 0.511] \\

\bottomrule
\end{tabular}
\end{table}

\begin{table}[t]
\centering
\small
\caption{Anytime held-out performance of LLaMEA on BBOB using the Devstral backbone. Cells report AOCC at five search-budget checkpoints ($20\%$, $40\%$, $60\%$, $80\%$, and $100\%$), evaluated on held-out instances across all 24 BBOB functions and dimensions $5$, $10$, and $20$. \emph{All} denotes the mean across the three dimensions. Main entries are medians across five search seeds, with interquartile ranges shown in brackets. Higher AOCC is better. (Part 2)}
\vspace{0.5em}
\label{tab:llamea_bbob_raw_results_devstral_p2}
\begin{tabular}{llcccc}
\toprule
Budget & Method & $d{=}5$ & $d{=}10$ & $d{=}20$ & All \\
\midrule
\multirow{14}{*}{60\%}
 & (1+1)-LLaMEA & 0.513 & 0.463 & 0.430 & 0.469 \\
 & & \scriptsize[0.476, 0.530] & \scriptsize[0.436, 0.470] & \scriptsize[0.400, 0.438] & \scriptsize[0.437, 0.480] \\
 & (10+10)-LLaMEA & 0.552 & 0.494 & 0.433 & 0.488 \\
 & & \scriptsize[0.539, 0.557] & \scriptsize[0.476, 0.496] & \scriptsize[0.408, 0.437] & \scriptsize[0.483, 0.501] \\
 & Fixed-1 & 0.472 & 0.426 & 0.395 & 0.430 \\
 & & \scriptsize[0.441, 0.513] & \scriptsize[0.409, 0.460] & \scriptsize[0.391, 0.417] & \scriptsize[0.414, 0.463] \\
 & Fixed-25\% & 0.561 & 0.473 & 0.396 & 0.484 \\
 & & \scriptsize[0.554, 0.580] & \scriptsize[0.468, 0.516] & \scriptsize[0.386, 0.492] & \scriptsize[0.476, 0.519] \\
 & Fixed-50\% & 0.555 & 0.463 & 0.417 & 0.469 \\
 & & \scriptsize[0.538, 0.562] & \scriptsize[0.453, 0.478] & \scriptsize[0.381, 0.421] & \scriptsize[0.466, 0.489] \\
 & AdaEva-S & 0.554 & \textbf{0.525} & \textbf{0.500} & \textbf{0.526} \\
 & & \scriptsize[0.554, 0.558] & \scriptsize[0.521, 0.531] & \scriptsize[0.474, 0.512] & \scriptsize[0.524, 0.532] \\
 & AdaEva-R & \textbf{0.587} & 0.488 & 0.426 & 0.511 \\
 & & \scriptsize[0.581, 0.600] & \scriptsize[0.477, 0.499] & \scriptsize[0.416, 0.451] & \scriptsize[0.489, 0.511] \\
\cmidrule(lr){1-6}
\multirow{14}{*}{80\%}
 & (1+1)-LLaMEA & 0.497 & 0.458 & 0.424 & 0.460 \\
 & & \scriptsize[0.412, 0.529] & \scriptsize[0.368, 0.468] & \scriptsize[0.328, 0.436] & \scriptsize[0.370, 0.478] \\
 & (10+10)-LLaMEA & 0.552 & 0.494 & 0.433 & 0.488 \\
 & & \scriptsize[0.539, 0.563] & \scriptsize[0.478, 0.496] & \scriptsize[0.420, 0.437] & \scriptsize[0.487, 0.501] \\
 & Fixed-1 & 0.472 & 0.426 & 0.395 & 0.431 \\
 & & \scriptsize[0.442, 0.513] & \scriptsize[0.410, 0.460] & \scriptsize[0.393, 0.417] & \scriptsize[0.415, 0.463] \\
 & Fixed-25\% & 0.559 & 0.462 & 0.385 & 0.475 \\
 & & \scriptsize[0.555, 0.570] & \scriptsize[0.460, 0.515] & \scriptsize[0.384, 0.479] & \scriptsize[0.468, 0.516] \\
 & Fixed-50\% & 0.557 & 0.461 & 0.411 & 0.470 \\
 & & \scriptsize[0.540, 0.562] & \scriptsize[0.457, 0.477] & \scriptsize[0.391, 0.416] & \scriptsize[0.468, 0.488] \\
 & AdaEva-S & 0.555 & \textbf{0.533} & \textbf{0.501} & \textbf{0.535} \\
 & & \scriptsize[0.554, 0.556] & \scriptsize[0.525, 0.533] & \scriptsize[0.498, 0.517] & \scriptsize[0.526, 0.536] \\
 & AdaEva-R & \textbf{0.591} & 0.488 & 0.427 & 0.502 \\
 & & \scriptsize[0.581, 0.604] & \scriptsize[0.476, 0.489] & \scriptsize[0.409, 0.435] & \scriptsize[0.488, 0.511] \\
\cmidrule(lr){1-6}
\multirow{14}{*}{100\%}
 & (1+1)-LLaMEA & 0.497 & 0.458 & 0.424 & 0.460 \\
 & & \scriptsize[0.412, 0.529] & \scriptsize[0.368, 0.468] & \scriptsize[0.328, 0.436] & \scriptsize[0.370, 0.478] \\
 & (10+10)-LLaMEA & 0.547 & 0.476 & 0.420 & 0.485 \\
 & & \scriptsize[0.541, 0.561] & \scriptsize[0.474, 0.491] & \scriptsize[0.408, 0.433] & \scriptsize[0.483, 0.504] \\
 & Fixed-1 & 0.468 & 0.425 & 0.395 & 0.429 \\
 & & \scriptsize[0.434, 0.514] & \scriptsize[0.407, 0.460] & \scriptsize[0.392, 0.417] & \scriptsize[0.411, 0.463] \\
 & Fixed-25\% & 0.563 & 0.464 & 0.386 & 0.474 \\
 & & \scriptsize[0.560, 0.568] & \scriptsize[0.460, 0.516] & \scriptsize[0.384, 0.492] & \scriptsize[0.470, 0.519] \\
 & Fixed-50\% & 0.558 & 0.464 & 0.411 & 0.470 \\
 & & \scriptsize[0.539, 0.562] & \scriptsize[0.457, 0.476] & \scriptsize[0.391, 0.415] & \scriptsize[0.467, 0.488] \\
 & AdaEva-S & 0.559 & \textbf{0.531} & \textbf{0.500} & \textbf{0.531} \\
 & & \scriptsize[0.555, 0.562] & \scriptsize[0.528, 0.535] & \scriptsize[0.494, 0.515] & \scriptsize[0.529, 0.533] \\
 & AdaEva-R & \textbf{0.591} & 0.489 & 0.426 & 0.504 \\
 & & \scriptsize[0.588, 0.600] & \scriptsize[0.476, 0.489] & \scriptsize[0.411, 0.435] & \scriptsize[0.489, 0.512] \\
\bottomrule
\end{tabular}
\end{table}

\begin{table}[t]
\centering
\small
\caption{Anytime held-out performance of LLaMEA on BBOB using the Qwen2.5-Coder backbone. Cells report AOCC at five search-budget checkpoints ($20\%$, $40\%$, $60\%$, $80\%$, and $100\%$), evaluated on held-out instances across all 24 BBOB functions and dimensions $5$, $10$, and $20$. \emph{All} denotes the mean across the three dimensions. Main entries are medians across five search seeds, with interquartile ranges shown in brackets. Higher AOCC is better. (Part 1)}
\vspace{0.5em}
\label{tab:llamea_bbob_qwen_p1}
\begin{tabular}{llcccc}
\toprule
Budget & Method & $d{=}5$ & $d{=}10$ & $d{=}20$ & All \\
\midrule
\multirow{12}{*}{20\%}
 & Fixed-1 & 0.499 & 0.425 & 0.365 & 0.435 \\
 & & \scriptsize[0.000, 0.515] & \scriptsize[0.000, 0.432] & \scriptsize[0.000, 0.385] & \scriptsize[0.000, 0.439] \\
 & Fixed-25\% & 0.545 & 0.474 & 0.431 & 0.477 \\
 & & \scriptsize[0.526, 0.568] & \scriptsize[0.453, 0.488] & \scriptsize[0.377, 0.434] & \scriptsize[0.473, 0.497] \\
 & Fixed-50\% & 0.544 & 0.451 & 0.404 & 0.473 \\
 & & \scriptsize[0.537, 0.624] & \scriptsize[0.451, 0.460] & \scriptsize[0.389, 0.415] & \scriptsize[0.462, 0.483] \\
 & Full & 0.548 & 0.480 & 0.420 & 0.482 \\
 & & \scriptsize[0.546, 0.563] & \scriptsize[0.465, 0.481] & \scriptsize[0.402, 0.425] & \scriptsize[0.470, 0.489] \\
 & AdaEva-S & 0.554 & \textbf{0.516} & \textbf{0.475} & 0.517 \\
 & & \scriptsize[0.547, 0.558] & \scriptsize[0.496, 0.520] & \scriptsize[0.456, 0.487] & \scriptsize[0.498, 0.518] \\
 & AdaEva-R & \textbf{0.583} & 0.506 & 0.462 & \textbf{0.520} \\
 & & \scriptsize[0.572, 0.598] & \scriptsize[0.492, 0.513] & \scriptsize[0.437, 0.465] & \scriptsize[0.500, 0.524] \\
\cmidrule(lr){1-6}
\multirow{12}{*}{40\%}
 & Fixed-1 & 0.499 & 0.425 & 0.365 & 0.435 \\
 & & \scriptsize[0.000, 0.515] & \scriptsize[0.000, 0.432] & \scriptsize[0.000, 0.385] & \scriptsize[0.000, 0.439] \\
 & Fixed-25\% & 0.552 & 0.490 & 0.437 & 0.503 \\
 & & \scriptsize[0.548, 0.594] & \scriptsize[0.456, 0.513] & \scriptsize[0.430, 0.448] & \scriptsize[0.476, 0.506] \\
 & Fixed-50\% & 0.553 & 0.463 & 0.414 & 0.479 \\
 & & \scriptsize[0.546, 0.614] & \scriptsize[0.458, 0.468] & \scriptsize[0.385, 0.423] & \scriptsize[0.477, 0.486] \\
 & Full & 0.565 & 0.481 & 0.420 & 0.487 \\
 & & \scriptsize[0.556, 0.572] & \scriptsize[0.471, 0.497] & \scriptsize[0.415, 0.426] & \scriptsize[0.482, 0.506] \\
 & AdaEva-S & 0.559 & 0.517 & \textbf{0.475} & 0.518 \\
 & & \scriptsize[0.555, 0.566] & \scriptsize[0.510, 0.519] & \scriptsize[0.468, 0.483] & \scriptsize[0.516, 0.518] \\
 & AdaEva-R & \textbf{0.589} & \textbf{0.518} & 0.472 & \textbf{0.530} \\
 & & \scriptsize[0.585, 0.623] & \scriptsize[0.515, 0.526] & \scriptsize[0.463, 0.479] & \scriptsize[0.525, 0.540] \\
\bottomrule
\end{tabular}
\end{table}

\begin{table}[t]
\centering
\small
\caption{Anytime held-out performance of LLaMEA on BBOB using the Qwen2.5-Coder backbone. Cells report AOCC at five search-budget checkpoints ($20\%$, $40\%$, $60\%$, $80\%$, and $100\%$), evaluated on held-out instances across all 24 BBOB functions and dimensions $5$, $10$, and $20$. \emph{All} denotes the mean across the three dimensions. Main entries are medians across five search seeds, with interquartile ranges shown in brackets. Higher AOCC is better. (Part 2)}
\vspace{0.5em}
\label{tab:llamea_bbob_qwen_p1}
\begin{tabular}{llcccc}
\toprule
Budget & Method & $d{=}5$ & $d{=}10$ & $d{=}20$ & All \\
\midrule
\multirow{12}{*}{60\%}
 & Fixed-1 & 0.499 & 0.425 & 0.365 & 0.435 \\
 & & \scriptsize[0.000, 0.515] & \scriptsize[0.000, 0.432] & \scriptsize[0.000, 0.385] & \scriptsize[0.000, 0.439] \\
 & Fixed-25\% & 0.560 & 0.456 & 0.445 & 0.476 \\
 & & \scriptsize[0.433, 0.561] & \scriptsize[0.341, 0.496] & \scriptsize[0.428, 0.455] & \scriptsize[0.400, 0.504] \\
 & Fixed-50\% & 0.557 & 0.465 & 0.409 & 0.487 \\
 & & \scriptsize[0.419, 0.596] & \scriptsize[0.453, 0.485] & \scriptsize[0.376, 0.437] & \scriptsize[0.411, 0.493] \\
 & Full & 0.580 & 0.488 & 0.419 & 0.495 \\
 & & \scriptsize[0.577, 0.581] & \scriptsize[0.474, 0.500] & \scriptsize[0.411, 0.456] & \scriptsize[0.494, 0.512] \\
 & AdaEva-S & 0.565 & 0.519 & 0.466 & 0.519 \\
 & & \scriptsize[0.563, 0.567] & \scriptsize[0.514, 0.521] & \scriptsize[0.444, 0.491] & \scriptsize[0.518, 0.523] \\
 & AdaEva-R & \textbf{0.609} & \textbf{0.523} & \textbf{0.479} & \textbf{0.533} \\
 & & \scriptsize[0.601, 0.619] & \scriptsize[0.523, 0.529] & \scriptsize[0.474, 0.488] & \scriptsize[0.532, 0.537] \\
\cmidrule(lr){1-6}
\multirow{12}{*}{80\%}
 & Fixed-1 & 0.499 & 0.425 & 0.365 & 0.435 \\
 & & \scriptsize[0.000, 0.515] & \scriptsize[0.000, 0.432] & \scriptsize[0.000, 0.385] & \scriptsize[0.000, 0.439] \\
 & Fixed-25\% & 0.560 & 0.509 & 0.448 & 0.504 \\
 & & \scriptsize[0.560, 0.596] & \scriptsize[0.498, 0.513] & \scriptsize[0.431, 0.453] & \scriptsize[0.503, 0.513] \\
 & Fixed-50\% & \textbf{0.608} & 0.487 & 0.440 & 0.512 \\
 & & \scriptsize[0.592, 0.628] & \scriptsize[0.485, 0.516] & \scriptsize[0.437, 0.468] & \scriptsize[0.493, 0.526] \\
 & Full & 0.594 & 0.510 & 0.448 & 0.520 \\
 & & \scriptsize[0.580, 0.598] & \scriptsize[0.499, 0.512] & \scriptsize[0.444, 0.455] & \scriptsize[0.519, 0.527] \\
 & AdaEva-S & 0.563 & 0.517 & \textbf{0.486} & 0.524 \\
 & & \scriptsize[0.558, 0.571] & \scriptsize[0.517, 0.524] & \scriptsize[0.463, 0.497] & \scriptsize[0.524, 0.526] \\
 & AdaEva-R & 0.600 & \textbf{0.525} & 0.482 & \textbf{0.534} \\
 & & \scriptsize[0.600, 0.621] & \scriptsize[0.523, 0.533] & \scriptsize[0.465, 0.485] & \scriptsize[0.529, 0.537] \\
\cmidrule(lr){1-6}
\multirow{12}{*}{100\%}
 & Fixed-1 & 0.499 & 0.425 & 0.365 & 0.435 \\
 & & \scriptsize[0.000, 0.515] & \scriptsize[0.000, 0.432] & \scriptsize[0.000, 0.385] & \scriptsize[0.000, 0.439] \\
 & Fixed-25\% & 0.560 & 0.509 & 0.448 & 0.504 \\
 & & \scriptsize[0.560, 0.596] & \scriptsize[0.498, 0.513] & \scriptsize[0.431, 0.452] & \scriptsize[0.503, 0.508] \\
 & Fixed-50\% & 0.610 & 0.497 & 0.441 & 0.513 \\
 & & \scriptsize[0.568, 0.626] & \scriptsize[0.489, 0.526] & \scriptsize[0.439, 0.462] & \scriptsize[0.502, 0.515] \\
 & Full & 0.599 & 0.510 & 0.458 & 0.527 \\
 & & \scriptsize[0.580, 0.615] & \scriptsize[0.499, 0.515] & \scriptsize[0.437, 0.459] & \scriptsize[0.524, 0.527] \\
 & AdaEva-S & 0.572 & 0.519 & \textbf{0.484} & 0.526 \\
 & & \scriptsize[0.562, 0.573] & \scriptsize[0.519, 0.522] & \scriptsize[0.467, 0.499] & \scriptsize[0.525, 0.530] \\
 & AdaEva-R & \textbf{0.616} & \textbf{0.531} & 0.479 & \textbf{0.538} \\
 & & \scriptsize[0.605, 0.618] & \scriptsize[0.527, 0.535] & \scriptsize[0.477, 0.485] & \scriptsize[0.534, 0.546] \\
\bottomrule
\end{tabular}
\end{table}

\begin{table}[t]
\centering
\small
\caption{Anytime held-out performance of EoH on TSPLIB using the Gemini-3.1-Flash-Lite backbone. Cells report the median optimality gap across five search seeds at five search-budget checkpoints ($20\%$, $40\%$, $60\%$, $80\%$, and $100\%$), with the interquartile range shown beneath. Results are grouped by instance size: $\leq 100$ (12 instances), $101$--$150$ (11 instances), and $>150$ (6 extrapolation instances). Lower values are better, and the best median within each budget block and instance-size group is shown in \textbf{bold}. (Part 1)}
\vspace{0.5em}
\label{tab:eoh_tsp_gemini}
\begin{tabular}{llccc}
\toprule
& & $\leq 100$ (in-distribution) & $101$--$150$ & $>150$ (extrapolation) \\
\cmidrule(lr){3-3}\cmidrule(lr){4-4}\cmidrule(lr){5-5}
Budget & Method & 12 instances & 11 instances & 6 instances \\
\midrule
\multirow{12}{*}{20\%}
 & Fixed-1 & 0.024 & 0.287 & 0.493 \\
 & & \scriptsize[0.004, 0.403] & \scriptsize[0.015, 0.404] & \scriptsize[0.107, 0.694] \\
 & Fixed-25\% & \textbf{0.020} & \textbf{0.037} & \textbf{0.338} \\
 & & \scriptsize[0.004, 0.403] & \scriptsize[0.002, 0.174] & \scriptsize[0.057, 0.635] \\
 & Fixed-50\% & 0.172 & 0.372 & 0.597 \\
 & & \scriptsize[0.028, 0.405] & \scriptsize[0.097, 0.681] & \scriptsize[0.385, 1.248] \\
 & Full & 0.608 & 1.513 & 1.612 \\
 & & \scriptsize[0.239, 0.951] & \scriptsize[0.871, 1.880] & \scriptsize[1.040, 3.635] \\
 & AdaEva-S & 0.170 & 0.372 & 0.632 \\
 & & \scriptsize[0.028, 0.409] & \scriptsize[0.052, 0.492] & \scriptsize[0.384, 0.930] \\
 & AdaEva-R & 0.024 & 0.157 & 0.360 \\
 & & \scriptsize[0.004, 0.403] & \scriptsize[0.015, 0.404] & \scriptsize[0.230, 0.553] \\
\cmidrule(lr){1-5}
\multirow{12}{*}{40\%}
 & Fixed-1 & 0.103 & 0.287 & 0.493 \\
 & & \scriptsize[0.008, 0.403] & \scriptsize[0.015, 0.378] & \scriptsize[0.107, 0.694] \\
 & Fixed-25\% & \textbf{0.024} & \textbf{0.028} & \textbf{0.253} \\
 & & \scriptsize[0.004, 0.403] & \scriptsize[0.002, 0.231] & \scriptsize[0.060, 0.502] \\
 & Fixed-50\% & 0.128 & 0.523 & 0.905 \\
 & & \scriptsize[0.026, 0.493] & \scriptsize[0.092, 1.247] & \scriptsize[0.642, 1.324] \\
 & Full & 0.123 & 0.450 & 0.826 \\
 & & \scriptsize[0.014, 0.403] & \scriptsize[0.039, 0.684] & \scriptsize[0.343, 1.211] \\
 & AdaEva-S & \textbf{0.024} & \textbf{0.028} & 0.331 \\
 & & \scriptsize[0.004, 0.403] & \scriptsize[0.002, 0.291] & \scriptsize[0.084, 0.345] \\
 & AdaEva-R & \textbf{0.024} & \textbf{0.028} & 0.279 \\
 & & \scriptsize[0.004, 0.403] & \scriptsize[-0.001, 0.218] & \scriptsize[0.068, 0.345] \\
\bottomrule
\end{tabular}
\end{table}

\begin{table}[t]
\centering
\small
\caption{Anytime held-out performance of EoH on TSPLIB using the Gemini-3.1-Flash-Lite backbone. Cells report the median optimality gap across five search seeds at five search-budget checkpoints ($20\%$, $40\%$, $60\%$, $80\%$, and $100\%$), with the interquartile range shown beneath. Results are grouped by instance size: $\leq 100$ (12 instances), $101$--$150$ (11 instances), and $>150$ (6 extrapolation instances). Lower values are better, and the best median within each budget block and instance-size group is shown in \textbf{bold}. (Part 2)}
\vspace{0.5em}
\label{tab:eoh_tsp_gemini}
\begin{tabular}{llccc}
\toprule
& & $\leq 100$ (in-distribution) & $101$--$150$ & $>150$ (extrapolation) \\
\cmidrule(lr){3-3}\cmidrule(lr){4-4}\cmidrule(lr){5-5}
Budget & Method & 12 instances & 11 instances & 6 instances \\
\midrule
\multirow{12}{*}{60\%}
 & Fixed-1 & 0.052 & 0.247 & 0.400 \\
 & & \scriptsize[0.008, 0.403] & \scriptsize[0.015, 0.338] & \scriptsize[0.101, 0.655] \\
 & Fixed-25\% & 0.024 & 0.028 & 0.333 \\
 & & \scriptsize[0.004, 0.403] & \scriptsize[0.002, 0.236] & \scriptsize[0.060, 0.542] \\
 & Fixed-50\% & 0.103 & 0.107 & 0.623 \\
 & & \scriptsize[0.008, 0.403] & \scriptsize[0.015, 0.316] & \scriptsize[0.365, 1.116] \\
 & Full & 0.214 & 0.328 & 0.479 \\
 & & \scriptsize[0.014, 0.516] & \scriptsize[0.136, 0.759] & \scriptsize[0.259, 1.195] \\
 & AdaEva-S & \textbf{0.012} & 0.014 & 0.241 \\
 & & \scriptsize[0.003, 0.403] & \scriptsize[0.002, 0.066] & \scriptsize[0.050, 0.401] \\
 & AdaEva-R & 0.020 & \textbf{0.012} & \textbf{0.239} \\
 & & \scriptsize[0.004, 0.403] & \scriptsize[0.002, 0.112] & \scriptsize[0.053, 0.426] \\
\cmidrule(lr){1-5}
\multirow{12}{*}{80\%}
 & Fixed-1 & 0.134 & 0.247 & 0.400 \\
 & & \scriptsize[0.014, 0.403] & \scriptsize[0.015, 0.338] & \scriptsize[0.101, 0.655] \\
 & Fixed-25\% & \textbf{0.024} & 0.028 & 0.414 \\
 & & \scriptsize[0.004, 0.403] & \scriptsize[0.002, 0.111] & \scriptsize[0.100, 0.451] \\
 & Fixed-50\% & \textbf{0.024} & 0.244 & 0.493 \\
 & & \scriptsize[0.004, 0.405] & \scriptsize[0.015, 0.406] & \scriptsize[0.225, 0.791] \\
 & Full & 0.103 & 0.372 & 0.641 \\
 & & \scriptsize[0.008, 0.419] & \scriptsize[0.137, 0.502] & \scriptsize[0.259, 1.277] \\
 & AdaEva-S & \textbf{0.024} & \textbf{0.012} & \textbf{0.260} \\
 & & \scriptsize[0.004, 0.403] & \scriptsize[0.000, 0.063] & \scriptsize[0.049, 0.378] \\
 & AdaEva-R & \textbf{0.024} & 0.056 & 0.323 \\
 & & \scriptsize[0.004, 0.403] & \scriptsize[0.002, 0.255] & \scriptsize[0.192, 0.602] \\
\cmidrule(lr){1-5}
\multirow{12}{*}{100\%}
 & Fixed-1 & 0.134 & 0.287 & 0.400 \\
 & & \scriptsize[0.014, 0.403] & \scriptsize[0.015, 0.339] & \scriptsize[0.101, 0.655] \\
 & Fixed-25\% & 0.024 & 0.028 & 0.333 \\
 & & \scriptsize[0.004, 0.403] & \scriptsize[0.002, 0.106] & \scriptsize[0.059, 0.816] \\
 & Fixed-50\% & 0.024 & 0.033 & 0.400 \\
 & & \scriptsize[0.004, 0.403] & \scriptsize[0.007, 0.274] & \scriptsize[0.214, 0.571] \\
 & Full & \textbf{0.020} & 0.028 & 0.348 \\
 & & \scriptsize[0.004, 0.403] & \scriptsize[0.002, 0.214] & \scriptsize[0.083, 0.418] \\
 & AdaEva-S & 0.024 & \textbf{0.012} & \textbf{0.304} \\
 & & \scriptsize[0.004, 0.403] & \scriptsize[0.000, 0.095] & \scriptsize[0.049, 0.524] \\
 & AdaEva-R & 0.024 & 0.028 & 0.328 \\
 & & \scriptsize[0.004, 0.403] & \scriptsize[0.002, 0.112] & \scriptsize[0.070, 0.444] \\
\bottomrule
\end{tabular}
\end{table}

\clearpage
\section{Partial Evaluation and Framework Feedback}
\label{app:partial_eval_feedback}

This appendix describes how AdaEva interacts with framework feedback under partial evaluation. In the main experiments, each framework retains its native feedback mechanism; among the frameworks considered, only LLaMEA directly exposes measured costs in the prompt, and therefore requires a prompt modification for the common-evidence condition. EoH and HiFo-Prompt use evaluation results in their search-side mechanisms but do not include fitness values directly in the prompt (see~\Cref{app:frameworks}). We provide the native and modified LLaMEA prompts below, with the latter explicitly indicating the partial-evaluation setting and reporting costs on a common evaluation block.

\begin{tcolorbox}[redpromptbox, title=Plain LLaMEA Prompt (Original)]
\begin{itemize}[leftmargin=1.5em, itemsep=4pt]
    \item \textbf{System \& instance Specification}: Informs the LLM of its role as an algorithm designer synthesizing Python heuristic functions for the BBOB benchmark suite.
    \item \textbf{Uncalibrated Population Context}: Lists all current population candidates with their uncalibrated average scores:
    \begin{itemize}[leftmargin=1.2em, label={--}]
        \item Format: \texttt{<Algorithm\_Name>: <Description> (Score: <Mean\_AOCC>)}
    \end{itemize}
    \item \textbf{Selected Parent \& Code Context}: Displays the description, Python source code, execution feedback, and error messages (if any) of the target parent solution selected for mutation.
    \item \textbf{Mutation Directive \& Output Format}: Injects a randomly selected mutation operator directive and specifies the required Python function output structure.
\end{itemize}
\end{tcolorbox}


\begin{tcolorbox}[
    redpromptbox,
    title=Common-Evidence LLaMEA Prompt (Partial Evaluation),
    breakable
]
\begin{itemize}[leftmargin=1.5em, itemsep=4pt]
    \item \textbf{System \& instance Specification}: Informs the LLM of its role as an algorithm designer synthesizing Python heuristic functions for the BBOB benchmark suite.
    \item \textbf{\color{mistralorange}[ADD-ON 1] Shared instance Context Header}: States explicitly that all non-eliminated population members were evaluated on the \textbf{exact same set of shared (instance, seed) instances}:
    \begin{itemize}[leftmargin=1.2em, label={--}]
        \item Header: \textit{``The non-eliminated algorithms below were evaluated on the SAME $N$ shared (instance, seed) instances [$f_1i_1, f_2i_1, \dots$], so their scores are directly comparable (eliminated candidates are marked and score -inf).''}
    \end{itemize}
    \item \textbf{\color{mistralorange}[ADD-ON 2] Per-BBOB-Class Performance Breakdown}: Exposes a granular breakdown of AOCC performance across the 5 standard BBOB problem classes (\textit{Separable, Low Conditioning, High Conditioning, Multi-Modal Adequate Structure, Multi-Modal Weak Structure}):
    \begin{itemize}[leftmargin=1.2em, label={--}]
        \item Format: \texttt{<Name>: <Desc> (Score: <Fit>, per-class AOCC [<sep>, <low\_cond>, <high\_cond>, <mm\_struct>, <mm\_weak>])}
    \end{itemize}
    \item \textbf{\color{mistralorange}[ADD-ON 3] Explicit Race Status \& Elimination Tagging}: Clearly demarcates eliminated or crashed candidates so the LLM is not misled by obsolete or invalid code:
    \begin{itemize}[leftmargin=1.2em, label={--}]
        \item Eliminated: \texttt{<Name>: <Desc> (Score: -inf, ELIMINATED by the race --- kept only for diversity, not comparable)}
        \item Crashed: \texttt{<Name>: <Desc> (Score: -inf, FAILED/crashed and eliminated --- kept only for diversity)}
    \end{itemize}
    \item \textbf{Selected Parent \& Code Context}: Displays the description, Python source code, execution feedback, and error messages (if any) of the target parent solution selected for mutation.
    \item \textbf{Mutation Directive \& Output Format}: Injects a randomly selected mutation operator directive and specifies the required Python function output structure.
\end{itemize}
\end{tcolorbox}

\section{Sensitivity to the AdaEva Configuration}
\label{app:sensitivity}

\paragraph{Protocol.} We vary one parameter at a time while holding all others at their default values, so that each ablated configuration differs from the default of its own instantiation in exactly one entry and interactions between parameters are not covered. The sweeps use the Devstral backbone on the three EoH benchmarks and on LLaMEA/BBOB, with five seeds per configuration drawn from the same seed set as the main experiments. For each setting, configuration, and seed, we use the incumbent recorded at each of five budget checkpoints ($20\%$, $40\%$, $60\%$, $80\%$, and $100\%$ of the total evaluation budget) and scored on the full search pool. These values are taken directly from the same validation-trajectory data used to produce the corresponding validation curves.

Let $v_{k,i}(c)$ denote the resulting value of configuration $c$ at checkpoint $k$ and seed $i$, and let $c_0$ denote the default configuration of the same AdaEva instantiation. Because configurations and their defaults share the same seed set and therefore the same initial population for each paired seed, differences are computed within seed. We first verify that the seed identifiers match between $c$ and $c_0$; unmatched seeds are excluded from the comparison, and configurations with fewer than four matched seeds are flagged rather than silently averaged. For each checkpoint $k$ and matched seed $i$, we compute the percent difference relative to the default:
\begin{equation}
\label{eq:sensitivity_diff}
d_{k,i}(c)
=
\frac{v_{k,i}(c)-v_{k,i}(c_0)}
     {v_{k,i}(c_0)} \times 100\%,
\end{equation}
for minimization objectives. For BBOB, where higher AOCC is better, the sign is reversed so that positive values consistently indicate worse performance:
\begin{equation}
\label{eq:sensitivity}
\Delta(c)
=
\frac{1}{5}
\sum_{k}
\operatorname{median}_{i}\,
d_{k,i}(c),
\end{equation}
where the median is taken over matched seeds at each checkpoint. Thus, $\Delta(c)>0$ indicates that configuration $c$ performs worse than its default, while $\Delta(c)<0$ indicates an improvement. The default configuration has $\Delta(c_0)=0$ by construction.

\paragraph{Discussion.} The sensitivity results show that the default configurations are generally stable on the three EoH benchmarks, with most changes producing deviations of less than $1\%$ in incumbent quality. The largest effects on CO occur for AdaEva-R on OBP, where replacing Friedman with a paired $t$-test increases the deviation to $+2.39\%$, and for AdaEva-S when increasing $\underline{n}_0$ to $10$, which gives a $+1.41\%$ deviation. Other changes are comparatively small, and several configurations are slightly better than the default on individual benchmarks, indicating that the defaults are not uniformly optimal for every setting.

The sensitivity is substantially larger on LLaMEA/BBOB. For AdaEva-R, changing $\underline{n}_0$ from its default of $5$ to either $1$ or $10$ increases the deviation to $+21.37\%$ and $+32.29\%$, respectively, while increasing the evaluation step $\Delta$ to $3$ gives $+28.92\%$. The statistical and population-management choices also matter: replacing Friedman with a paired $t$-test, replacing Conover with Nemenyi, disabling population refill, and disabling stochastic evaluation result in deviations of $+14.93\%$, $+21.26\%$, $+19.48\%$, and $+21.16\%$, respectively. For AdaEva-S, the effects are smaller, with $\underline{n}_0=10$ changing performance by $-2.01\%$ and $\eta=1.5$ by $-7.74\%$ on BBOB. Finally, replacing the native LLaMEA prompt with the partial-evaluation prompt increases the deviation by $+14.99\%$, providing no evidence that the modified feedback is beneficial in this setting.

Overall, the results indicate that the sensitivity of AdaEva depends on both the instantiation and the benchmark. The CO results suggest that the default parameters are relatively robust to moderate changes, whereas BBOB is more sensitive to how AdaEva-R accumulates and interprets evaluation evidence. This supports using fixed, prescribed defaults rather than tuning the configuration separately for each benchmark, while also showing that the choice of racing parameters can materially affect performance in some settings.

\begin{table}[t]
\centering
\caption{Configuration sensitivity of AdaEva. Each cell is the percent change in the incumbent's full-pool quality under the ablated configuration relative to the default within the same seed, averaged across budget checkpoints ($\Delta(c)$); positive values indicate worse performance than the default. One factor is varied at a time, so each configuration differs from its default in exactly one parameter. $\underline{n}_0$ is varied separately for AdaEva-S and AdaEva-R, while $\eta$ is specific to AdaEva-S and the racing parameters are specific to AdaEva-R. The prompt rows correspond to the LLaMEA feedback ablation. Defaults are marked $\dagger$, and -- denotes a configuration with no usable run.}
\label{tab:sensitivity}
\vspace{0.5em}
\begin{tabular}{llrrrr}
\toprule
Parameter & Value & EoH/OBP & EoH/TSP & EoH/FSSP & LLaMEA/BBOB \\
\midrule
\multirow{4}{*}{$\underline{n}_0$ (AdaEva-R)}  & 1 & $-0.22$ & $+0.02$ & $0.00$ & $+21.37$ \\
 & 5 & $0^{\dagger}$ & $0^{\dagger}$ & $0^{\dagger}$ & $0^{\dagger}$ \\
 & 10 & $+0.84$ & $+0.05$ & $-0.03$ & $+32.29$ \\
\cmidrule(lr){1-6}
\multirow{2}{*}{$\Delta$} & 1 & $0^{\dagger}$ & $0^{\dagger}$ & $0^{\dagger}$ & $0^{\dagger}$ \\
 & 3 & $+1.05$ & $+0.02$ & $-0.01$ & $+28.92$ \\
\cmidrule(lr){1-6}
\multirow{3}{*}{$\underline{n}_0$ (AdaEva-S)} & 1 & $-0.03$ & $-0.08$ & $+0.02$ & $+2.83$ \\
 & 5 & $0^{\dagger}$ & $0^{\dagger}$ & $0^{\dagger}$ & $0^{\dagger}$ \\
 & 10 & $+1.41$ & $-0.17$ & $-0.02$ & $-2.01$ \\
\midrule
\multirow{3}{*}{$\alpha$} & 0.05 & $0^{\dagger}$ & $0^{\dagger}$ & $0^{\dagger}$ & $0^{\dagger}$ \\
 & 0.01 & $+0.48$ & $+0.08$ & $-0.02$ & $+24.23$ \\
 & 0.005 & $+0.61$ & $+0.03$ & $-0.04$ & $-0.13$ \\
\cmidrule(lr){1-6}
\multirow{2}{*}{Statistical Test} & Friedman & $0^{\dagger}$ & $0^{\dagger}$ & $0^{\dagger}$ & $0^{\dagger}$ \\
 & $t$-test & $+2.39$ & $+0.04$ & $-0.06$ & $+14.93$ \\
\cmidrule(lr){1-6}
\multirow{2}{*}{Post-hoc Test} & Conover & $0^{\dagger}$ & $0^{\dagger}$ & $0^{\dagger}$ & $0^{\dagger}$ \\
 & Nemenyi & $+1.07$ & $+0.04$ & $-0.04$ & $+21.26$ \\
\cmidrule(lr){1-6}
\multirow{2}{*}{$\ell$} & 2 & $0^{\dagger}$ & $0^{\dagger}$ & $0^{\dagger}$ & $0^{\dagger}$ \\
 & 6 & $+0.96$ & $+0.01$ & $-0.02$ & $+4.81$ \\
\cmidrule(lr){1-6}
\multirow{2}{*}{$\eta$} & 1.25 & $0^{\dagger}$ & $0^{\dagger}$ & $0^{\dagger}$ & $0^{\dagger}$ \\
 & 1.5 & $+0.25$ & $+0.26$ & $-0.02$ & $-7.74$ \\
\midrule
\multirow{2}{*}{Prompt} & Original & -- & -- & -- & $0^{\dagger}$ \\
 & Partial Evaluation & -- & -- & -- & $+14.99$ \\
\midrule
\multirow{2}{*}{Population Refill} & True & $0^{\dagger}$ & $0^{\dagger}$ & $0^{\dagger}$ & $0^{\dagger}$ \\
 & False & $-0.39$ & $+0.03$ & $-0.01$ & $+19.48$ \\
\cmidrule(lr){1-6}
\multirow{2}{*}{Stochastic Evaluation} & True & $0^{\dagger}$ & $0^{\dagger}$ & $0^{\dagger}$ & $0^{\dagger}$ \\
 & False & $+0.41$ & $+0.09$ & $+0.03$ & $+21.16$ \\
\bottomrule
\end{tabular}
\end{table}

\end{document}

%% file: math_commands.tex
\usepackage{amsmath,amsfonts,bm}

\def\eqref#1{equation~\ref{#1}}

\def\1{\bm{1}}

\DeclareMathAlphabet{\mathsfit}{\encodingdefault}{\sfdefault}{m}{sl}
\SetMathAlphabet{\mathsfit}{bold}{\encodingdefault}{\sfdefault}{bx}{n}

